%% file: main.tex
\documentclass[10pt,journal,compsoc]{IEEEtran}

\usepackage{tikz}
\usepackage{comment}
\usepackage{pgfplots}
\usepackage{xcolor}
\usepackage{filecontents}
\usepackage{tablefootnote}
\usepackage{amsmath}
\usepackage{graphicx}
\usepackage{url}
\usepackage{hyperref}
\usepackage{subcaption}
\usepackage{amsfonts}
\usepackage{color, soul}
\usepackage{mathrsfs}
\usepackage{cleveref}
\usepackage{multirow}
\usepackage{float}
\usepackage{wrapfig}
\usepackage{amsthm}
\usepackage{bm}
\usepackage{bbm}
\usepackage{algorithm}
\usepackage{algorithmic}
\usepackage{colortbl}
\usepackage{enumitem}
\usepackage{color}
\usepackage{balance}
\usepackage{stfloats}
\usepackage{makecell}
\usepackage{dsfont}
\usepackage{amsmath}
\usepackage{amsthm}
\usepackage{amssymb}
\usepackage{amsfonts}
\usepackage{tikz}
\usepackage{forest}

\newtheorem{problem}{Problem}
  \usepackage{pifont}
\usepackage[utf8]{inputenc} % allow utf-8 input
\usepackage[T1]{fontenc}    % use 8-bit T1 fonts
\usepackage{hyperref}       % hyperlinks
\usepackage{url}            % simple URL typesetting
\usepackage{booktabs}       % professional-quality tables
\usepackage{amsfonts}       % blackboard math symbols
\usepackage{nicefrac}       % compact symbols for 1/2, etc.
\usepackage{microtype}      % microtypography
\usepackage{xcolor}         % colors

\def \st{\;\text{s.t.}\;}

\newcommand{\bX}{\mathbf{X}}

\newcommand{\bA}{\mathbf{A}}

\newcommand{\bh}{\mathbf{h}}

\newcommand{\bx}{\mathbf{x}}

\DeclareMathOperator*{\argmin}{argmin}

\usepackage{xcolor,pifont}

\newcommand*\colourcheck[1]{%
  \expandafter\newcommand\csname #1check\endcsname{\textcolor{#1}{\ding{52}}}%
}
\colourcheck{black}
\colourcheck{blue}
\colourcheck{green}
\colourcheck{red}

\newcommand*\colourwrong[1]{%
  \expandafter\newcommand\csname #1wrong\endcsname{\textcolor{#1}{\ding{55}}}%
}
\colourwrong{blue}
\colourwrong{green}
\colourwrong{red}

\ifCLASSOPTIONcompsoc
  \usepackage[nocompress]{cite}
\else
  \usepackage{cite}
\fi

\ifCLASSINFOpdf

\else

\fi

\begin{document}

\title{Safety in Graph Machine Learning: \\Threats and Safeguards \\  }

\author{Song Wang, Yushun Dong, Binchi Zhang, Zihan Chen, Xingbo Fu, Yinhan He, \\ Cong Shen, Chuxu Zhang, Nitesh V. Chawla, and Jundong Li% <-this % stops a space
\IEEEcompsocitemizethanks{
\IEEEcompsocthanksitem S. Wang, Y. Dong, B. Zhang, Z. Chen, X. Fu, Y. He, C. Shen, and J. Li are with the Department of Electrical and Computer Engineering, University of Virginia, Charlottesville, Virginia, USA.\protect\\
E-mails: \{sw3wv, yd6eb, epb6gw, brf3rx, xf3av, nee7ne, cong, jundong\}@virginia.edu
\IEEEcompsocthanksitem C. Zhang is with the Department of Computer Science, Brandeis University, Waltham, Massachusetts, USA.\protect\\
E-mail: chuxuzhang@brandeis.edu
\IEEEcompsocthanksitem N. V. Chawla is with the Department of Computer Science and Engineering, University of Notre Dame, Notre Dame, Indiana, USA.\protect\\
E-mail:  nchawla@nd.edu
}% <-this % stops a space
}

% The paper headers
\markboth{IEEE TRANSACTIONS ON KNOWLEDGE AND DATA ENGINEERING}%
{Shell \MakeLowercase{\textit{et al.}}: Bare Advanced Demo of IEEEtran.cls for IEEE Computer Society Journals}

\IEEEtitleabstractindextext{% demographic subgroups and items do not be so specific
\begin{abstract}

Graph Machine Learning (Graph ML) has witnessed substantial advancements in recent years. With their remarkable ability to process graph-structured data, Graph ML techniques have been extensively utilized across diverse applications, including critical domains like finance, healthcare, and transportation. Despite their societal benefits, recent research highlights significant safety concerns associated with the widespread use of Graph ML models. Lacking safety-focused designs, these models can produce unreliable predictions, demonstrate poor generalizability, and compromise data confidentiality. In high-stakes scenarios such as financial fraud detection, these vulnerabilities could jeopardize both individuals and society at large. Therefore, it is imperative to prioritize the development of safety-oriented Graph ML models to mitigate these risks and enhance public confidence in their applications.
In this survey paper, we explore three critical aspects vital for enhancing safety in Graph ML: reliability, generalizability, and confidentiality. We categorize and analyze threats to each aspect under three headings: model threats, data threats, and attack threats. This novel taxonomy guides our review of effective strategies to protect against these threats. Our systematic review lays a groundwork for future research aimed at developing practical, safety-centered Graph ML models. Furthermore, we highlight the significance of safe Graph ML practices and suggest promising avenues for further investigation in this crucial area.

\end{abstract}

% Note that keywords are not normally used for peerreview papers.
}

\maketitle

\IEEEdisplaynontitleabstractindextext

\IEEEpeerreviewmaketitle

\linespread{1}
\selectfont

\section{Introduction}
\label{sec:introduction}

\label{intro}

In recent years, graph-structured data has become increasingly prevalent across a wide range of real-world applications, including drug discovery \cite{bongini2021molecular}, traffic forecasting \cite{jiang2022graph}, and disease diagnosis \cite{li2020graph}. Within these domains, Graph Machine Learning (Graph ML) plays a pivotal role in modeling this data and executing graph-based predictive tasks \cite{kipf2017semi,xu2018powerful}. However, as the scope of Graph ML applications expands, concerns about their underlying safety issues intensify \cite{dai2022comprehensive}. Inadequately addressing these issues can result in severe implications, particularly in critical decision-making scenarios \cite{zhang2022trustworthy}. For instance, in financial fraud detection, Graph ML models analyze transaction networks where nodes represent users and edges depict transactions \cite{suzumura2019towards}. Susceptibility to shifts in data distribution could erroneously flag legitimate transactions as fraudulent \cite{dai2022comprehensive}. Additionally, these models may also pose risks to user privacy \cite{orekondy2019knockoff}. Both these safety concerns significantly erode trust in the financial system.

Despite growing societal concerns \cite{sun2022adversarial,xu2023machine}, a comprehensive understanding of safety within Graph Machine Learning (Graph ML) is still emerging. This lack of understanding impedes researchers and practitioners from systematically identifying and addressing the fundamental safety concerns associated with Graph ML methods.
To narrow this gap, our survey seeks to resolve two critical questions: (1) What are the key aspects involved in the safety issues of Graph ML? (2) What specific types of threats might arise within each aspect, and how can they be effectively handled?
To address the first question, we introduce a novel taxonomy that facilitates a thorough categorization of safety issues in Graph ML. To answer the second question, we provide a systematic review of potential threats and their corresponding safeguards within each aspect identified in our taxonomy.

We now delve into the first question, identifying three key aspects of safety issues in Graph ML: \emph{reliability}, \emph{generalizability}, and \emph{confidentiality}. We discuss each aspect separately below.
% We now elaborate on the first question. Specifically, we identify three key aspects of the safety issues in Graph ML, including \emph{reliability}, \emph{generalizability}, and \emph{confidentiality}.
% %
% Below we introduce them separately. 
\textit{(1) Reliability}:
Graph ML models often face challenges from low-quality training data, which may stem from data noise \cite{ding2019deep,xu2022contrastive} or malicious attacks \cite{dou2020enhancing,bojchevski2019adversarial}. Here, we define reliability as the model's ability to consistently produce high-quality outputs even when confronted with substandard input. Reliability becomes particularly critical when high-quality training data for Graph ML tasks is scarce \cite{bandyopadhyay2020outlier,bhardwaj2021adversarial}. For example, in drug discovery, Graph ML models predict the chemical properties of new compounds, modeled as molecular graphs~\cite{li2022out}. This includes assessing toxicity to eliminate drugs with severe side effects \cite{mervin2021uncertainty}. Due to the prohibitive costs associated with experimental validations, high-quality labels for training data are hard to come by. Therefore, maintaining model reliability in the face of low-quality data is essential to ensure accurate predictions \cite{jiang2021could}.
\textit{(2) Generalizability}: Generalizability refers to the ability of Graph ML models to perform consistently well across various scenarios, especially when there are changes in the underlying data distribution \cite{li2022out}. This is crucial in environments where new, unseen graph data frequently emerges \cite{wu2019domain}.
For instance, in pandemic prevention, accurate prediction of future infections is essential to allocate medical resources effectively \cite{zeroual2020deep}. Graph ML is extensively used to forecast confirmed cases using the topological data from different geographical areas (such as counties, states, or countries) interconnected by migration paths \cite{soriano2020impact}.
However, pandemics typically result in asynchronous outbreaks across different regions, leading to training and inference data for Graph ML models being derived from diverse areas \cite{panagopoulos2021transfer}. The lack of generalizability in these models can cause the predicted infection rates to deviate significantly from actual scenarios, potentially resulting in the misallocation of medical resources.
\textit{(3) Confidentiality}: 
This aspect focuses on safeguarding the integrity of Graph ML models and the privacy of the sensitive data they process \cite{defazio2019adversarial, cong2022grapheditor}. Confidentiality is particularly crucial in Graph ML applications that handle personal data \cite{zhang2022trustworthy}. For instance, Graph ML is extensively utilized in managing Electronic Health Records (EHR) for tasks such as disease diagnosis and treatment prediction \cite{lu2020decentralized}. In these applications, various types of information—such as patient names and diagnosis outcomes—are interconnected to form heterogeneous graphs \cite{liu2020heterogeneous}. However, a significant challenge to confidentiality in Graph ML arises from the prevalent use of message-passing mechanisms, which facilitate the flow of information between nodes \cite{dai2022comprehensive, bayram2021federated, said2023survey}. This process can inadvertently allow sensitive data to reach malicious nodes, leading to potential disclosures of private information to unauthorized entities, thereby compromising patient privacy \cite{fernandez2013security}.

We then address the second question by identifying three types of threats that compromise the safety of Graph ML, applicable across all previously discussed aspects. Specifically, the threats include (1) \textit{model threats}, which stem from the inherent learning mechanisms designed for graph structures, such as message-passing, prevalent in most Graph ML models; (2) \textit{data threats}, originating from the complex interrelationships among nodes within the graph topology; and (3) \textit{attack threats}, which occur due to vulnerabilities in Graph ML models when exposed to adversarial attacks. In this survey, we aim to provide a comprehensive understanding of the origins of each threat type and offer a systematic overview of current safeguarding techniques to mitigate these risks.

\begin{table*}[!t]\centering
\caption{
Comparison of our survey with other studies in the field, focusing on each identified threat within the three key safety aspects.
}\vspace{-0.0in}
\renewcommand{\arraystretch}{1.25}
\setlength{\tabcolsep}{8.2pt}
\scalebox{1}{
  \setlength{\aboverulesep}{0pt}
  \setlength{\belowrulesep}{0pt}
  \begin{tabular}{c|cc|ccc|ccc|ccc}
    \midrule[1pt]
\multirow{2}{*}{\textbf{Surveys}}&\multicolumn{2}{c|}{\textbf{Scope}}&\multicolumn{3}{c|}{\textbf{Reliability}}&\multicolumn{3}{c|}{\textbf{Generalizability}}&\multicolumn{3}{c}{\textbf{Confidentiality}}\\\cline{2-12}
&Safety&Graph ML&Model&Data&Attack&Model&Data&Attack&Model&Data&Attack\\\hline
Li et al.~\cite{li2023trustworthy}&\redcheck&\rule[0.5ex]{0.6em}{1pt}&\rule[0.5ex]{0.6em}{1pt}&\redcheck&\redcheck&\redcheck&\rule[0.5ex]{0.6em}{1pt}&\redcheck&\redcheck&\redcheck&\redcheck\\
Liu et al.~\cite{liu2022trustworthy}&\redcheck&\rule[0.5ex]{0.6em}{1pt}&\rule[0.5ex]{0.6em}{1pt}&\rule[0.5ex]{0.6em}{1pt}&\redcheck&\rule[0.5ex]{0.6em}{1pt}&\rule[0.5ex]{0.6em}{1pt}&\redcheck&\redcheck&\redcheck&\rule[0.5ex]{0.6em}{1pt}
\\\hline
Jin et al.~\cite{jin2021adversarial}&\redcheck&\redcheck&\rule[0.5ex]{0.6em}{1pt}&\rule[0.5ex]{0.6em}{1pt}&\redcheck&\rule[0.5ex]{0.6em}{1pt}&\rule[0.5ex]{0.6em}{1pt}&\redcheck&\rule[0.5ex]{0.6em}{1pt}&\rule[0.5ex]{0.6em}{1pt}&\rule[0.5ex]{0.6em}{1pt}\\
Li et al.~\cite{li2022out}&\rule[0.5ex]{0.6em}{1pt}&\redcheck&\rule[0.5ex]{0.6em}{1pt}&\rule[0.5ex]{0.6em}{1pt}&\rule[0.5ex]{0.6em}{1pt}&\redcheck&\redcheck&\rule[0.5ex]{0.6em}{1pt}&\rule[0.5ex]{0.6em}{1pt}&\rule[0.5ex]{0.6em}{1pt}&\rule[0.5ex]{0.6em}{1pt}\\\hline
Wu et al.~\cite{wu2022survey}&\redcheck&\redcheck&\rule[0.5ex]{0.6em}{1pt}&\rule[0.5ex]{0.6em}{1pt}&\redcheck&\rule[0.5ex]{0.6em}{1pt}&\redcheck&\redcheck&\redcheck&\redcheck&\rule[0.5ex]{0.6em}{1pt}\\
Dai et al.~\cite{dai2022comprehensive}&\redcheck&\redcheck&\rule[0.5ex]{0.6em}{1pt}&\rule[0.5ex]{0.6em}{1pt}&\redcheck&\rule[0.5ex]{0.6em}{1pt}&\rule[0.5ex]{0.6em}{1pt}&\redcheck&\redcheck&\redcheck&\redcheck \\
Zhang et al.~\cite{zhang2022trustworthy}&\redcheck&\redcheck&\rule[0.5ex]{0.6em}{1pt}&\rule[0.5ex]{0.6em}{1pt}&\redcheck&\rule[0.5ex]{0.6em}{1pt}&\rule[0.5ex]{0.6em}{1pt}&\redcheck&\redcheck&\redcheck&\redcheck  \\\hline
Ours&\redcheck&\redcheck&\redcheck&\redcheck&\redcheck &\redcheck&\redcheck&\redcheck &\redcheck&\redcheck&\redcheck \\
    \midrule[1pt]
  \end{tabular}}
  \vspace{-.0in}
  \label{tab:comparison}
\end{table*}

\noindent\textbf{Contributions.} In this survey, we conduct a comprehensive investigation of safety issues in Graph ML and organize the key aspects of Graph ML safety in a structured framework (as illustrated in Fig.~\ref{fig:taxonomy-techniques}). Additionally, we outline three types of safety threats for each aspect and delve into specific safeguard techniques designed to address these issues within the context of Graph ML. In conclusion, our work makes three significant contributions to the research community.
\begin{itemize}[leftmargin=0.35cm]
\item \textbf{Novel Taxonomy for Graph ML Safety}: We introduce a novel taxonomy to categorize the safety issues in Graph ML, detailing three core aspects: reliability, generalizability, and confidentiality.

\item \textbf{Comprehensive Overview of Threats and Safeguards}: We identify three distinct types of threats common to all safety aspects. Building on this foundation, we further explore safeguard techniques tailored to address each specific threat.

\item \textbf{Future Research Potential in Graph ML Safety}: We explore open challenges and emerging opportunities in enhancing safety within the field of Graph ML, aiming to inspire future research initiatives.
\end{itemize}

\noindent\textbf{Comparison to other surveys.} 
There have been a few surveys investigating different types of safety issues and countermeasures for Graph ML~\cite{sun2022adversarial, xu2020adversarial, jin2021adversarial,wu2022survey,zhang2023survey,li2022out, gui2022good}. 
However, these surveys often lack a comprehensive understanding of safety in Graph ML.
Other related surveys focus on a different, yet related topic: trustworthiness~\cite{dai2022comprehensive,zhang2022trustworthy}. Most of these works overwhelmingly focus on adversarial attacks, while the safety issues closely tailored for Graph ML and graph data are usually neglected.
Different from the above works, in this survey, we not only provide a systematic review of a broader notion of safety for Graph ML with the help of our newly proposed taxonomy, but also elaborate on different types of threats, along with the safeguard techniques to properly handle them. We provide a detailed comparison of the differences in Table~\ref{tab:comparison}.

\section{Preliminaries of Graph ML}

\input{category}

\subsection{Notations}
Throughout this survey, we represent a graph as $\mathcal{G}=(\mathcal{V},\mathcal{E})$, with $\mathcal{V}=\{v_i\}_{i=1}^N$ and $\mathcal{E}\subseteq\mathcal{V}\times\mathcal{V}$ being the node set and the edge set, respectively. $N=|\mathcal{V}|$ is the number of nodes in $\mathcal{G}$, i.e., the size of graph $\mathcal{G}$. $e_{ij}=(v_i,v_j)\in\mathcal{E}$ is an edge between $v_i$ and $v_j$. We additionally consider the (optional) node features $\bX\in\mathbb{R}^{N\times d}$, where $\bx_i\in\mathbb{R}^d$ is the $i$-th row vector of $\bX$ and also a $d$-dimensional feature vector of node $v_i$. We also represent the adjacency matrix of $\mathcal{G}$ as $\bA\in\mathbb{R}^{N\times N}$, where $\bA_{ij}=1$ if there exists an edge between $v_i$ and $v_j$, and $\bA_{ij}=0$, otherwise. For simplicity, we focus on undirected graphs in this survey. We use bold uppercase characters to denote matrices (e.g., $\bX$) and bold lowercase characters to denote vectors (e.g., $\bx$). In Table~\ref{tab:notation}, we provide detailed descriptions for the notations used in this survey.

\subsection{Graph ML Methodologies}
% Graph ML methodologies have been broadly investigated for analyzing graph-structured data.
% % Initially, random-walk methods (e.g., DeepWalk~\cite{perozzi2014deepwalk} and Node2Vec~\cite{grover2016node2vec}) have been applied to learn node representations. 
% In recent years, Graph Neural Networks (GNNs) have gained prominence over traditional alternatives (e.g., DeepWalk~\cite{perozzi2014deepwalk} and Node2Vec~\cite{grover2016node2vec}) due to their superiority in a variety of tasks~\cite{kipf2017semi, xu2018powerful, hamilton2017inductive}. GNNs are generally based on the message-passing scheme. Particularly, the essence of the message-passing lies in updating node representations by iteratively aggregating and processing information from adjacent nodes. The process of representation learning via this message-passing scheme, in the $k$-th layer of GNNs is given as

Graph ML methodologies have been extensively explored for analyzing graph-structured data. Recently, Graph Neural Networks (GNNs) have emerged as a superior alternative to traditional methods such as DeepWalk \cite{perozzi2014deepwalk} and Node2Vec \cite{grover2016node2vec}, demonstrating their effectiveness across various tasks \cite{kipf2017semi, xu2018powerful, hamilton2017inductive}. GNNs primarily operate on a message-passing scheme, where the core mechanism involves updating node representations by iteratively aggregating and processing information from neighboring nodes. This representation learning process within the $k$-th layer of GNNs is described as follows
\begin{equation}
    \mathbf{m}_v^{(k)}= \text{Aggregate}(\{\bh_u^{(k-1)}|u\in\mathcal{N}_v),
    \end{equation}
    \begin{equation}
    \bh_v^{(k)}= \text{Update}(\mathbf{m}_v^{(k)}, \bh_v^{(k-1)}),
\end{equation}
where $\bh_v^{(k)}$ is the representation of node $v$ after the $k$-th GNN layer, and $\mathcal{N}_v$ denotes
the neighboring node set of $v$. The learned node representations can be applied to various graph-related tasks, e.g., node classification, link prediction, and graph classification. For instance, in node classification, a classifier is typically trained to map node representations to specific labels \cite{kipf2017semi}. In graph classification, a read-out function, often implemented as averaging, is used to aggregate node embeddings into a single graph-level embedding. For a more detailed discussion on learning with graphs, refer to \cite{zhang2020deep}.
% A $k$-layer GNN thus learns the representation for each node by leveraging the information from its $k$-hop neighboring nodes.

\begin{table}[!t]\centering
\caption{
Key notations used in this paper.
}\vspace{-0.05in}
\renewcommand{\arraystretch}{1.2}
\setlength{\tabcolsep}{5.2pt}
\scalebox{1}{
  \setlength{\aboverulesep}{0pt}
  \setlength{\belowrulesep}{0pt}
  \begin{tabular}{l|l}
    \midrule[1pt]
\textbf{Notations}&\textbf{Description}\\\hline
$\mathcal{G}$& A given graph.\\
$\mathcal{V}$& The node set of $\mathcal{G}$.\\
$v\in\mathcal{V}$& A node in $\mathcal{V}$.\\
$\mathcal{N}_v$& The neighbor set of node $v$.\\
$\mathcal{E}$& The edge set of $\mathcal{G}$.\\
$e\in\mathcal{E}$ &An edge in $\mathcal{E}$.\\
$N$& The number of nodes in $\mathcal{G}$, i.e., $|\mathcal{V}|$.\\
$d$& The dimension size of a node feature.\\
$\bX\in\mathbb{R}^{N\times d}$& The node feature matrix.\\
$\bA\in\mathbb{R}^{N\times N}$& The adjacency matrix.\\
$d_h$& The dimension size of the hidden representations.\\
$\bh_v\in\mathbb{R}^{d_z}$& The hidden representation of node $v$. \\
$f(\cdot)$& A Graph ML model.\\
    \midrule[1pt]
  \end{tabular}}
  \vspace{-0.1in}
  \label{tab:notation}
\end{table}

\section{Reliability in Graph ML}

Due to the complex nature of graph data, acquiring high-quality training data is a significant challenge~\cite{ding2019deep, dou2020enhancing}. When models are trained on data of poor quality, their performance can be severely compromised~\cite{
xu2022contrastive}. We define reliability as the ability of a model to maintain consistent performance, even when trained on lower-quality data. The lack of reliability exposes models to the risks associated with noisy or tampered data, potentially leading to incorrect predictions. This issue is particularly critical in decision-making processes, where such inaccuracies could result in unacceptable outcomes \cite{dallachiesa2014node, hu2017embedding}. For example, in financial fraud detection, it is crucial to accurately identify fraudulent transactions. A model that lacks reliability might fail to detect fraudulent activities or falsely label legitimate transactions as fraudulent, resulting in financial losses. Thus, enhancing reliability in Graph ML models is vital for ensuring their safety and dependability in critical applications, ultimately reducing the risk of erroneous outcomes \cite{pal2019bayesian, ng2018bayesian}.

Reliability in Graph ML can be compromised from several angles. First, the model's inherent limitations in processing uncertain data can lead to overconfident predictions in unfamiliar situations \cite{hasanzadeh2020bayesian}. Unlike image or text data, graph data involves complex interactions between nodes, and uncertainties in these connections can impact model predictions \cite{zhuang2022uncertainty}. Second, anomalies in training data, such as nodes and graphs that deviate significantly from typical distributions, pose a significant threat to model performance and, consequently, reliability \cite{ding2019deep, bandyopadhyay2020outlier}. Since Graph ML models aggregate information from neighboring nodes, anomalous nodes can influence the representations learned across the graph, affecting overall model reliability. Third, poisoning attacks involve the insertion of maliciously crafted data into training sets to compromise model reliability \cite{li2020adversarial}. Within the graph topology, attackers can manipulate a few nodes to adversely affect other distant nodes, making these attacks particularly difficult to detect \cite{jin2021adversarial}.

Below, we discuss a range of safeguard techniques aimed at enhancing model reliability against these threats.
% , in order to ensure safer Graph ML applications. %even in the face of diverse and challenging application scenarios.

\subsection{Model Threat: Uncertainty} 
% Uncertainty refers to the variance within the model predictions, indicating the degree to which a model is unsure about its predictions~\cite{pal2019bayesian,ng2018bayesian}.
% In Graph ML models, uncertainty in the predictions could severely impact model reliability, especially in high-stakes applications~\cite{hendrycks2021unsolved}. For example, in drug discovery using Graph ML models, when a compound is predicted to have a high probability of being effective against a certain disease but with high uncertainty, further experiments should be conducted to verify the compound's efficacy~\cite{jiang2021could,mervin2021uncertainty}.
Uncertainty in model predictions refers to the variance within these predictions, highlighting the extent to which a model is unsure about its outcomes \cite{pal2019bayesian, ng2018bayesian}. In Graph ML models, such uncertainty can significantly undermine reliability, particularly in high-stakes applications \cite{hendrycks2021unsolved}. For instance, in drug discovery, if a Graph ML model predicts that a compound is likely effective against a particular disease but with high uncertainty, additional experiments are necessary to confirm the compound’s efficacy \cite{jiang2021could, mervin2021uncertainty}.
%%%measuring model uncertainty against each prediction could assist in the detection process by identifying predictions that require further investigation.
%
Uncertainty quantification \cite{zhang2019bayesian} serves as a crucial technique to estimate the confidence level of each prediction in Graph ML models. This approach enables the identification of predictions that require further validation, thus enhancing the overall reliability of the model. Specifically, uncertainty quantification can address two main types of uncertainty: \emph{Aleatory Uncertainty}, which arises from noisy input data \cite{hu2017embedding}, and \emph{Epistemic Uncertainty}, which results from imperfect model parameters \cite{hasanzadeh2020bayesian}.

In the following sections, we explore research efforts that utilize uncertainty quantification to develop more reliable Graph ML models.

%we introduce existing efforts dedicated to building more reliable Graph ML models through uncertainty quantification. 

\subsubsection{Aleatory Uncertainty}

Aleatory uncertainty pertains to the inherent randomness in data caused by noisy or imprecise measurements. In Graph ML, aleatory uncertainty primarily originates from two sources: the node feature matrix $\mathbf{X}$ and the adjacency matrix $\mathbf{A}$. For instance, in a protein-protein interaction (PPI) network \cite{raman2010construction}, each node represents a protein characterized by a feature vector detailing its chemical properties, while each edge denotes the interaction between two proteins. Both the chemical properties (i.e., node features) and the interactions (i.e., edge connections) are subject to measurement errors and inherent variability, contributing to uncertainties in both graph node features and topology \cite{yin2023omg}. When Graph ML models employing message-passing mechanisms, such as GCN~\cite{kipf2017semi}, are used on these uncertain nodes and edges, this uncertainty can propagate through the network, significantly affecting the learning process.

Currently, most studies addressing aleatory uncertainty concentrate on the uncertainty inherent in graph structure, typically resulting from imprecise measurements of the graph's configuration. The prevalent approach to modeling this type of uncertainty is to treat the input graph as an uncertain graph \cite{dallachiesa2014node, hu2017embedding}. In such an uncertain graph, each pair of nodes is assigned a probability value that indicates the likelihood of an edge existing between them. For example, {Bayesian GCN} \cite{zhang2019bayesian} investigates uncertain graphs derived from a parametric family of random graphs and estimates the prediction distribution of a Graph ML model using the expression $p(\mathbf{Z}|\mathbf{Y}, \mathbf{X}, \mathcal{G}_{obs})$, which is defined as follows:
\begin{equation}\label{equ:bayesian_gnn}
\begin{aligned}
    \int p(\mathbf{Z}|\mathbf{W}, \mathcal{G}, \mathbf{X})p(\mathbf{W}|\mathbf{Y}, \mathbf{X}, \mathcal{G})p(\mathcal{G}|\lambda)p(\lambda|\mathcal{G}_{obs})d\mathbf{W}d\mathcal{G}d\lambda.
\end{aligned}
\end{equation}
In this context, $\mathbf{Z}$ represents the output of the Graph ML model, such as the predicted node label distribution matrix. $\mathbf{Y}$ refers to the node labels from the training data. $\mathcal{G}_{obs}$ denotes the observed input graph, while $\mathcal{G}$ represents the uncertain graph. $\mathbf{W}$ represents the weights of a Bayesian GCN, and $\lambda$ indicates the parameters governing the probability distribution of the uncertain graph

However, it is important to note that the Bayesian GCN relies heavily on the probability distribution of the uncertain graph, where $p(\lambda|\mathcal{G}_{obs})$ estimates its parameters, $\lambda$. Subsequently, $p(\mathcal{G}|\lambda)$ describes the distribution of the uncertain graph conditioned on its parameters $\lambda$. This model primarily captures information about the graph structure and neglects node features. To address this, {BGCN} \cite{pal2019bayesian} introduces a non-parametric version of the Bayesian GCN, which integrates node feature information into the GNN predictions. Both the Bayesian GCN and its non-parametric counterpart involve high computational complexity due to the necessity of Monte Carlo sampling for estimating the probability distribution of the uncertain graphs. To construct a more efficient uncertainty-aware Graph ML model, {GGP} \cite{ng2018bayesian} utilizes a data-efficient Gaussian process-based Bayesian approach, which speeds up the training process while still delivering satisfactory performance.

\subsubsection{Episdemic Uncertainty}\label{subsec:epi_uncertainty}

Epistemic uncertainty, also known as systematic uncertainty, arises from knowledge gaps in what could theoretically be known but remains unknown in practice \cite{der2009aleatory}. This type of uncertainty often stems from a model's omission of certain factors, leading to an incomplete representation of the underlying process. In Graph ML, epistemic uncertainty specifically refers to the uncertainty associated with the model parameters. To measure the epistemic uncertainty of Graph ML models, current studies typically fix the input graph's adjacency matrix (to control aleatory uncertainty) and estimate the model's prediction error using Bayesian approaches \cite{zhang2017estimating, hasanzadeh2020bayesian}. As a classic example, {GCN-BBGDC} \cite{hasanzadeh2020bayesian} quantifies GNN epistemic uncertainty through Patch Accuracy versus Patch Uncertainty (PAvPU). PAvPU assesses the ratio of accurate certain predictions to inaccurate uncertain predictions, with higher PAvPU values indicating that that certain predictions are more accurate while inaccurate predictions are more uncertain. Additionally, the Mean Prediction Interval Width (MPIW) serves as another measure of GNN epistemic uncertainty, calculated as $\text{MPIW}=\frac{1}{N}\sum\nolimits_{i=1}^{N}(U_i-L_i)$, where $N$ is the number of nodes in the input graph, and $L_i$ and $U_i$ are the endpoints of the confidence interval for each node's prediction. However, PAvPU provides an overall model uncertainty value rather than individual node uncertainty. Addressing this, {STZINB-GNN} \cite{zhuang2022uncertainty} calculates a 90\% confidence interval for each node's prediction, offering a more granular view of epistemic uncertainty. Similarly, {CF-GNN} \cite{huang2023uncertainty} applies conformal prediction to estimate epistemic uncertainty at the node level.

\begin{figure}\center
\includegraphics[width=0.4\textwidth] {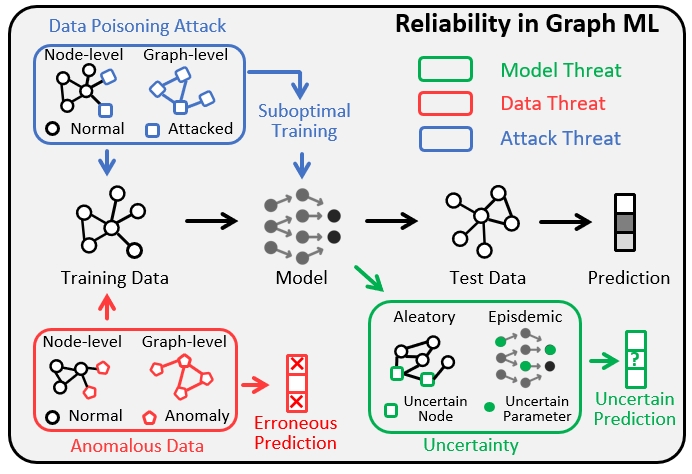}
\vspace{-.1in}
\caption{Safety threats regarding reliability in Graph ML.}
\vspace*{-.15in}
\label{fig:reliability}
\end{figure}

\subsection{Data Threat: Anomalies}

Anomalies in graph data, such as abnormal nodes and edges with noisy and outlying patterns that significantly deviate from the normal distribution \cite{ding2019deep, bandyopadhyay2020outlier}, pose considerable challenges. For example, fake users on social networks intended to spread misinformation represent such anomalies \cite{zhu2020mixedad}. In Graph ML models, the presence of these data anomalies can severely compromise model reliability by impairing the learning process. Consider a scenario where a Graph ML model is trained to classify news articles on social networks as credible or not \cite{li2018ssdmv}. If the training dataset includes fake news articles not identified as anomalies, the model may erroneously classify these articles as credible. This can lead to a failure in accurately detecting new instances of fake news, thereby undermining the model's reliability \cite{ma2021comprehensive}.

Recently, a variety of graph anomaly detection (GAD) techniques have been developed to mitigate the negative impact of anomalies \cite{xu2022contrastive,ma2021comprehensive,zhao2023using,dou2021user}. By identifying and removing anomalous patterns, we can enhance the reliability of Graph ML models. Unlike anomalies in Euclidean spaces, such as images, anomalies in graphs present unique challenges due to the complex interactions among nodes \cite{zhang2019dane}. For example, in social networks, fake users may connect with numerous real users to blend in and expand their influence \cite{feng2022twibot}. Detecting such anomalies necessitates a deep understanding of complex graph topologies and node relationships \cite{liu2021pick}. We will now discuss various effective methodologies for detecting anomalies in graph data. We categorize GAD techniques into two types based on the nature of the anomalies: techniques for detecting \textit{in-graph anomalies} and \textit{cross-graph anomalies}.

\subsubsection{In-Graph Anomalies}
% In-graph anomaly detection aims to identify anomalous graph components, including nodes, edges, and subgraphs. In this subsection, we primarily focus on node-level detection, given its prevalence in existing works. 
%We primarily focus on node-level anomaly detection as it is the majority of existing works.

In-graph anomaly detection focuses on identifying anomalous components within a graph, such as nodes, edges, and subgraphs. In this subsection, we concentrate on node-level detection, as it is the most commonly addressed aspect in existing research.

\vspace{0.02in}
\noindent\textbf{Reconstruction-based Methods.}
% Typically, anomalous nodes exhibit significant differences compared to other nodes~\cite{liu2021anomaly}, such as an attacker account within networks of commercial systems or an individual fraudulent user in social media networks~\cite{zhu2020mixedad}. Hence, a feasible strategy is to learn a model to reconstruct the information of each node and leverage the reconstruction error for detection. Intuitively, a node with a larger reconstruction error is more likely to be anomalous. Accordingly, the anomaly score $a$ of node $v$ is estimated as follows:
Typically, anomalous nodes in a network, such as attacker accounts in commercial systems or fraudulent users in social media networks, exhibit significant deviations compared to regular nodes \cite{liu2021anomaly, zhu2020mixedad}. A common strategy for detecting these anomalies involves training a model to reconstruct the information of each node and using the reconstruction error for anomaly detection~\cite{ding2021inductive}. Intuitively, a node with a higher reconstruction error is more likely to be anomalous. Consequently, the anomaly score $a$ of node $v$ is estimated as follows:
\begin{equation}
    a \propto \alpha \cdot d(\mathbf{h}_s,\hat{\mathbf{h}}_s)+\beta\cdot d(\mathbf{h}_x,\hat{\mathbf{h}}_x),
\end{equation}
where $\mathbf{h}_s$ and $\mathbf{h}_x$ represent the learned representations of node $v$ from its local structures and node features, respectively. The function $d(\cdot,\cdot)$ measures the distance between two inputs, while $\alpha$ and $\beta$ are hyperparameters that control the weight of these distances. With this model, anomaly scores are calculated for each node, identifying those with higher scores as potential anomalies \cite{ma2021comprehensive}.
Among various methods, Radar \cite{li2017radar} uses matrix decomposition to model the residuals of attribute information and its coherence with network information, using these residuals to detect anomalies. ANOMALOUS~\cite{peng2018anomalous} mitigates the impact of noisy and structurally irrelevant attributes via attribute selection and detects anomalies with residual analysis while focusing on representative attributes.
More recent approaches leverage Graph Neural Networks (GNNs) to derive comprehensive node representations for GAD. For example, DOMINANT \cite{ding2019deep} employs Graph Convolutional Networks (GCNs) to learn and then reconstruct graph structure and node features using separate decoders, also implemented as GCNs. 
Similarly, ResGCN \cite{pei2022resgcn} uses GCNs along with a linear classifier to model the node feature matrix. Like Radar, ResGCN calculates a residual matrix during the reconstruction process and uses these residuals to compute anomaly scores.

\vspace{0.02in}
\noindent\textbf{Contrastive Learning Methods.} 
Recent studies have highlighted that in reconstruction-based methods, the optimization objective of reconstruction is not inherently linked to anomaly detection. This is because the model may still learn to reconstruct anomalous patterns in the graph data \cite{xu2022contrastive}. As a response, newer GAD methods have begun incorporating contrastive learning \cite{oord2018representation}, which differentiates between various representations of nodes (referred to as contrastive pairs). The degree of similarity between nodes serves as an indicator of their anomalousness \cite{zheng2021generative}. {CoLA} \cite{liu2021anomaly} uses random walks to sample two subgraphs for each node as a contrastive pair and calculates the anomaly score based on the differences between the target node and the sampled subgraphs. However, CoLA's random sampling does not capitalize on prior knowledge. Addressing this, {CONAD} \cite{xu2022contrastive} integrates anomaly types identified by human experts to generate varied representations for the contrastive pair, using contrastive learning to distinguish these representations and compute anomaly scores based on their discrepancies. Still, these methods lack the ability to adaptively generate contrastive pairs for each node. To refine this process, {Sub-CR} \cite{zhang2022reconstruction} employs graph diffusion techniques to create contrastive pairs from both local and global perspectives.

\subsubsection{Cross-Graph Anomalies}

Cross-graph anomaly detection focuses on identifying individual graphs that display significant deviations compared to other graphs \cite{zhao2023using,dou2021user}. Initially, researchers approached the detection of cross-graph anomalies using traditional graph representation techniques, such as Graph2Vec \cite{narayanan2017graph2vec}. This approach transforms the task of cross-graph anomaly detection into a conventional anomaly detection problem with independent and identically distributed (i.i.d.) data \cite{lee2020anomaly}. Typically, these representation-based methods employ a two-stage strategy \cite{ma2021comprehensive}: first, graphs are encoded into a shared latent space using graph-level representation methods; then, an anomaly detection algorithm assesses the anomalousness of each graph. Despite their straightforward nature, these methods often suffer from suboptimal performance, as the representations they learn may not effectively capture the nuances necessary for detecting anomalies \cite{lee2021gawd}.

Recent advancements in cross-graph anomaly detection have leveraged GNNs to develop more sophisticated graph representations \cite{hu2022deep,zhu2022memory}. For instance, {UPFD} \cite{dou2021user} approaches the challenge of fake news detection by treating news articles as tree-like propagation graphs, where the root node represents the news article and the surrounding nodes signify users spreading the article. For each news graph, representations derived from both the text and graph structures are used to identify anomalies. Another notable development, {OCGIN} \cite{zhao2023using}, employs a one-class classification loss \cite{ruff2018deep} to create an end-to-end framework for cross-graph anomaly detection. In this model, the embedding for each graph is generated by mean-pooling the embeddings of its nodes, with graphs deemed anomalous if they fall outside the defined hypersphere.

\subsection{Attack Threat: Data Poisoning}

Recent studies have highlighted how adversarial attacks on training data, often referred to as data poisoning attacks, can significantly compromise the training phase of Graph ML models \cite{zugner2018adversarial,dai2018adversarial,jin2021adversarial,sun2018data}. These attacks involve subtle yet deliberate modifications to nodes and edges within the graph data, aimed at undermining the training process \cite{sun2019node,zhang2023adversarial}. Such data poisoning attacks pose a considerable threat to the reliability of Graph ML models by deteriorating their performance. For example, spammers on social networks can act as adversarial samples \cite{feng2022twibot}. Unlike normal users, spammers may disseminate fake and sensitive content, hijack trending topics, and misuse mention functions \cite{li2018ssdmv}. The actions of spammers can significantly disrupt the utility of Graph ML models applied to normal users due to the extensive connections within the graph \cite{akoglu2015graph}. This leads to less accurate predictions from the trained models, posing a severe threat to their reliability.

Although defensive strategies for data types like images have been extensively explored, the inherent complexity of graph structures significantly complicates the task of defending against attacks \cite{chen2022graphfraudster}. In graphs, perturbations targeting even a single node can impact a broad array of connected nodes due to the interconnected nature of the data \cite{liu2022gradients}. To address these unique challenges, various defensive techniques have been developed recently to protect graphs by identifying and mitigating the effects of data poisoning attacks \cite{jin2019power}. Below, we discuss recent advancements in attack mechanisms and defensive strategies against data poisoning, with further details provided in Table~\ref{tab:poisoning_attack}.

\subsubsection{Data Poisoning Attack}
%We first explore the various methodologies for executing data poisoning attacks in Graph ML models.
%
We first formulate the problem of data poisoning attacks on graphs as a bilevel optimization task:
\begin{problem}\label{prob:attack}
Given a graph $\mathcal{G}=(\mathcal{V},\mathcal{E})$, let $f:\mathcal{G}\rightarrow Z$ be a Graph ML model, where $Z$ denotes the node embeddings.
The data poisoning attacks on graphs can be formulated as
\begin{equation}
\begin{aligned}
&\max_{\mathcal{G}^{\prime}\in\mathcal{F}}\mathcal{L}_{atk}(\mathcal{G}^{\prime},f^{*}),\quad
\st \;f^{*}=\argmin_f\mathcal{L}(\mathcal{G}^{\prime},f),
\end{aligned}
\end{equation}
where $\mathcal{L}_{atk}$ denotes the attacker's objective function and $\mathcal{F}$ denotes the feasible space of poisoned graphs.
\end{problem}

Data poisoning attacks on graphs can be categorized based on the attacker's knowledge into three types: \textit{white-box} attacks \cite{xu2019topology}, \textit{gray-box} attacks \cite{kang2023deceptive}, and \textit{black-box} attacks \cite{sun2018data}. In white-box attacks, the attacker has complete access to the victim's model and data, including model parameters, training data, and ground truth labels \cite{bhardwaj2021adversarial,bhardwaj2021poisoning,zhang2021projective}. While white-box attacks can be highly effective on Graph ML models, meeting the conditions for such attacks is often impractical in real-world scenarios \cite{jin2021adversarial}.
Gray-box attacks provide the attacker with limited knowledge of the victim's model and data, excluding access to the exact model parameters \cite{zhang2023adversarial}. In these cases, attackers typically use a surrogate Graph ML model to predict the behavior of the victim model.
Black-box attacks further reduce the attacker's knowledge; the attacker only accesses the input graph data and the corresponding model output \cite{li2020adversarial}. Any details about model parameters, ground truth labels, or intermediate results are inaccessible \cite{chang2020restricted, wan2021adversarial}.
These attacks can be aimed at various graph-related tasks, such as node classification and link prediction, with the attacker's objectives varying across different scenarios.

% traditional tasks and advanced tasks, where traditional tasks include community detection and shallow node embedding, and advanced tasks contain node/graph classification, link prediction, and trustworthiness.
% Next, we discuss these goals separately.

\begin{table}[!t]
\centering
\caption{Different categories of methods for poisoning attacks for Graph ML.}
\vspace{-0.05in}
\setlength{\tabcolsep}{3.9pt}
\renewcommand{\arraystretch}{1.2}
\scalebox{1}{
\setlength{\aboverulesep}{0pt}
\setlength{\belowrulesep}{0pt}
\begin{tabular}{c|c|c}
\toprule[1pt]
 \textbf{Target Task} & \textbf{Setting} & \textbf{References} \\
\midrule
Community Detection & Black-box & \cite{li2020adversarial} \\
\cline{1-3}
Network Embedding & Black-box & \cite{bojchevski2019adversarial} \\
\cline{1-3}
\multirow{1}{*}{Node/Graph} & White-box & \cite{xu2019topology} \\
 \multirow{1}{*}{Classification}& Gray-box & \cite{zugner2018adversarial,zugner2020adversarial,liu2022gradients,wang2019attacking,sun2019node} \\
\cline{1-3}
 \multirow{2}{*}{Link Prediction} & White-box & \cite{bhardwaj2021adversarial,bhardwaj2021poisoning} \\
 & Black-box & \cite{sun2018data} \\
\cline{1-3}
 Trustworthiness & Gray-box & \cite{zhang2023adversarial,kang2023deceptive} \\
\bottomrule[1pt]
\end{tabular}}
\vspace{-.1in}
\label{tab:poisoning_attack}
\end{table}

% \noindent\textbf{Advanced Tasks on Graphs.}
\vspace{0.02in}
\noindent\textbf{Node Classification.}
For node classification tasks, \cite{xu2019topology} formulates the white-box poisoning attack as a bilevel optimization problem, where Projected Gradient Descent (PGD) \cite{nocedal1999numerical} is used to approximate a solution in the discrete graph topology space. Similarly, {Nettack} \cite{zugner2018adversarial}, a seminal method in the gray-box setting, substitutes the victim model with a simpler GCN surrogate. Nettack optimizes by greedily flipping edges to maximize the increase in loss over the targeted node, effectively navigating the discrete topology of graph data.
However, to mount a comprehensive attack on all nodes within a target graph, the greedy selection approach, while effective, becomes computationally intensive. {Metattack} \cite{zugner2020adversarial} addresses this by employing meta-learning \cite{finn2017model} to approximate the meta gradient of the loss function across all victim nodes in terms of the adjacency matrix, flipping the edge that maximizes the loss increment. Building on this, {AtkSE} \cite{liu2022gradients} incorporates strategies for edge discrete sampling and momentum gradient ensemble to mitigate the optimization instability caused by meta gradient inaccuracies.
Addressing the complexity of bilevel optimization, Wang et al. \cite{wang2019attacking} reformulate poisoning attacks against collective classification on graphs \cite{sen2008collective} as a constrained optimization problem, simplified through the use of Lagrange multipliers with PGD to resolve it within the discrete graph data domain. Beyond edge flipping, attackers may also introduce adversarial nodes into the graph to disrupt classification outcomes, as seen in NIPA \cite{sun2019node}, where attackers add new nodes without altering the existing graph structure. Additionally, {G-FairAttack} \cite{zhang2023adversarial} presents a gray-box approach that subtly undermines the group fairness of GNNs without noticeably affecting model utility. In a related effort, {FATE} \cite{kang2023deceptive} uses meta-learning to simultaneously target both group and individual fairness \cite{zugner2020adversarial}.

\vspace{0.02in}
\noindent\textbf{Link Prediction.}
{Opt-Attack} \cite{sun2018data} serves as a classic example of how the quality of unsupervised graph embeddings, such as DeepWalk \cite{perozzi2014deepwalk}, can be compromised in link prediction tasks. When the target is a knowledge graph, attackers may alter links or entities to disrupt the embeddings, consequently reducing link prediction accuracy. Bhardwaj et al. \cite{bhardwaj2021adversarial} employ the influence function \cite{koh2017understanding} to identify the knowledge triple that most significantly impacts the prediction of a target triple and replace it with a distant, irrelevant entity. Additionally, \cite{bhardwaj2021poisoning} targets knowledge graph embeddings by exploiting relational inference patterns. Specifically, the authors develop different scoring functions based on symmetry, inversion, and composition patterns to select the most impactful relations and knowledge entities for modification.

\vspace{0.02in}
\noindent\textbf{Other Tasks on Graphs.}
Data poisoning attacks are also performed on community detection~\cite{li2020adversarial} and network embedding~\cite{bojchevski2019adversarial}.
For community detection, CD-Attack~\cite{li2020adversarial} is a classic poisoning attack method in the black-box setting. This approach involves generating adversarial graphs that maximize the normalized cut while using a constrained VGAE model \cite{kipf2016variational} to obscure targeted entities from detection. In the realm of network embedding, Bojchevski et al. \cite{bojchevski2019adversarial} directly maximize the unsupervised loss function $\mathcal{L}_{atk}$ for DeepWalk \cite{perozzi2014deepwalk} in a black-box setting. Moreover, the effectiveness of these attacks can extend to other graph embedding methods such as node2vec \cite{grover2016node2vec} and GCN \cite{kipf2017semi}.

\subsubsection{Defense Against Data Poisoning Attacks}
% Now we delve into defense strategies designed to protect Graph ML models from data poisoning attacks.% We introduce a range of techniques developed to detect and mitigate the effects of adversarial perturbations on graph data.
We now explore defense strategies aimed at protecting Graph ML models from data poisoning attacks.

\vspace{0.02in}
\noindent \textbf{Poisoned Data Filtering.}
When faced with a potentially poisoned graph as input, various defense techniques aim to filter out the compromised elements before training begins. Entezari et al. \cite{entezari2020all} defend against the Nettack \cite{zugner2018adversarial} by replacing the poisoned graph with a low-rank approximation prior to the training process. NeuralSparse \cite{zheng2020robust} counters data poisoning attacks using a supervised graph sparsification module that removes task-irrelevant edges while learning graph representations. This approach utilizes a deep neural network to parameterize the edge selection process within the one-hop neighborhood, adhering to a fixed budget. In a similar vein, ProGNN \cite{jin2021adversarial} simultaneously learns a clean graph structure and a robust GNN to enhance the utility of the target GNN on unlabeled nodes of a poisoned graph. The defense's learning objective preserves several graph properties, including low rank, sparsity, and feature smoothness.
Given the significant computational demands of computing a low-rank approximation with $O(N^3)$ time complexity and $O(N^2)$ space complexity (where $N$ is the number of nodes), LRGNN \cite{xu2021speedup} decouples the adjacency matrix into low-rank and sparse components, focusing on minimizing the rank of the low-rank part while suppressing the sparse component. Aside from directly learning a clean graph, Zhang et al. \cite{zhang2021detection} propose methods to detect adversarial attacks on graphs. Specifically, they introduce two statistical tests to identify single-node poisoning and node-set corruption, employing Jensen-Shannon Divergence and Maximum Mean Discrepancy, respectively.

% Considering the high complexity of computing the low-rank approximation ($O(n^3)$ time complexity and $O(n^2)$ space complexity), \textit{LRGNN}~\cite{xu2021speedup} decouples the adjacency matrix into a low-rank and a sparse component and minimizes the rank of the low-rank part while suppressing the sparse part. %A sparse variant of LRGNN is formed to further reduce the memory cost. 
% Other than directly learning a clean graph, \cite{zhang2021detection} proposes to detect adversarial attacks on graphs. In particular, the authors propose two statistical tests for single-node poisoning and node-set corruption, based on Jensen-Shannon Divergence and Maximum Min Discrepancy, respectively.
\vspace{0.02in}
\noindent \textbf{Robust Graph Learning.}
Another approach to defending against data poisoning does not involve filtering out poisoned data; instead, it focuses on learning graph embeddings that are inherently robust against attacks. Tang et al. \cite{tang2020transferring} explore how to train a robust Graph ML model using a set of clean graphs by generating ground truth poisoned links on clean graphs to train an attack-aware Graph ML model. 
%This model incorporates a penalized aggregation mechanism that assigns low attention weights to the poisoned links. 
However, such supervision signals are often scarce in real-world scenarios. In the absence of explicit supervision signals, {SimPGCN} \cite{jin2021node} enhances robustness by maintaining feature similarity during the node aggregation process. SimPGCN designs a new aggregation mechanism that adaptively balances the information from the graph structure and node features. Further developments in robust Graph ML methods are explored in multiple studies such as \cite{liu2021graph,chang2021not,zhang2023chasing,yuan2024chasing}, each offering unique perspectives on improving resilience. %chen2021understanding,

\begin{figure}\center
\includegraphics[width=0.4\textwidth] {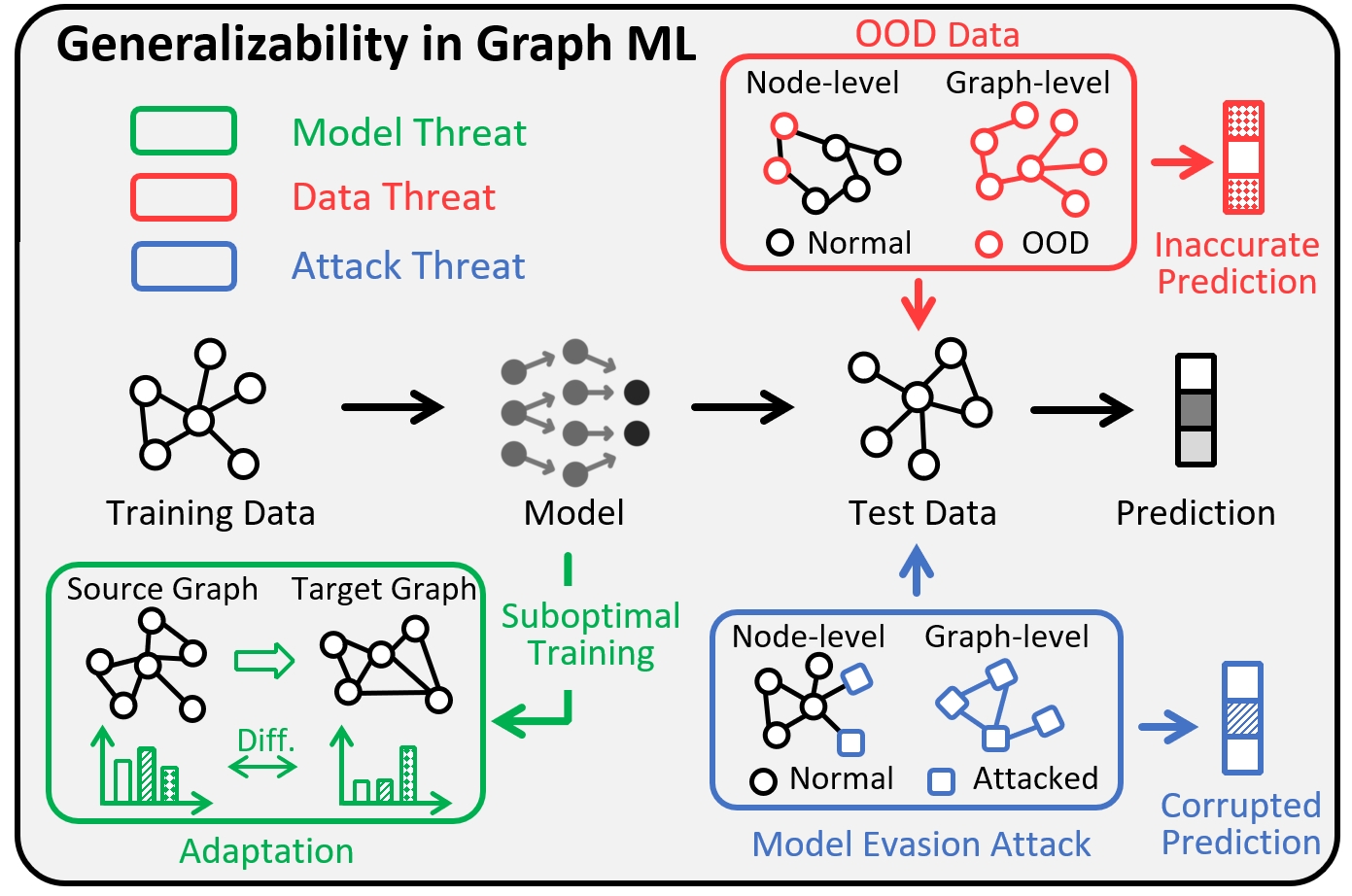}
\vspace*{-.1in}
\caption{Safety threats regarding generalizability in Graph ML.}
\vspace*{-.15in}
\label{fig:generalizability}
\end{figure}

\section{Generalizability in Graph ML}
% Most of the existing Graph ML models are developed based on the assumption that the data distribution during inference is consistent with that during trianing~\cite{li2022out}. However, this assumption is difficult to achieve in practice, due to the diverse and complex structures in graph data. As a result, the generalizability, i.e., maintaining model performance across different data distributions, becomes crucial in the safe deployment of Graph ML models.
Most existing Graph ML models are developed under the assumption that the data distribution during inference mirrors that during training \cite{li2022out}. However, this assumption often falls short in practice due to the complexity of graph data structures. Consequently, generalizability---defined as the ability to maintain consistent model performance across varying data distributions---becomes crucial for the safe deployment of Graph ML models.
%Due to the variety of graph data caused by the structures on graphs, Graph ML models are generally required to perform across various tasks. As a result, their performance is not guaranteed if not specifically trained for generalizability. 
%In this survey, we define generalizability in Graph ML as the capacity of maintaining model performance in different data distributions during inference.
%%%Particularly, generalizability is essential for Graph ML models to accurately transfer their learned knowledge to different applications with unseen or novel data~\cite{li2022out,dai2022comprehensive}. 
% For example, in drug discovery, Graph ML models are generally required to predict the toxicity of new drugs in different distributions from training data.
% Failure to generalize well to the new drugs can lead to untrustworthy predictions that jeopardize the development of medical treatments~\cite{guo2021few,chen2022learning}. 
For instance, in drug discovery, Graph ML models are typically tasked with predicting the toxicity of new drugs that may differ significantly from the training data distributions. A failure to generalize effectively to these new drugs can result in unreliable predictions, potentially compromising the development of medical treatments \cite{guo2021few,chen2022learning}.
%in decision-making applications where critical decisions could result in financial losses or affect human welfare~\cite{hendrycks2021unsolved}. 
%As such, enhancing generalizability is paramount for the safe and effective application of Graph ML technologies.

% Despite the importance of generalizability in Graph ML safety, different threats pose challenges to achieving model generalizability.
Despite the critical importance of generalizability for ensuring the safety of Graph ML models, various threats challenge the achievement of consistent model generalizability.
%
% First, the model threat to generalizability stems from a model's inability in its design when the model may not adequately adapt to the target (unlabeled) distributions that are available during training~\cite{zhu2021transfer, mallick2021transfer}. In the presence of data in different distributions during training, the model design to enhance generalizability becomes crucial for the broader applications of Graph ML models~\cite{mahapatra2022unsupervised}. 
First, the threat to generalizability from the model arises due to inherent limitations in its design, where the model may not effectively adapt to the target (unlabeled) distributions encountered during training \cite{zhu2021transfer, mallick2021transfer}. When training involves data from various distributions, designing models to enhance generalizability is crucial for the wide-ranging applications of Graph ML models \cite{mahapatra2022unsupervised}.
%This could result in overfitting to training data and lacking the flexibility for broader applications~\cite{mahapatra2022unsupervised}. 
%This threat is further exaggerated by the inherent complexity of graph data, characterized by its structured nature that involves nodes and edges.
%
% Second, the data threat arises from encountering unseen data distributions not covered during training, which can hinder the performance of Graph ML models in new scenarios~\cite{li2022learning}. As the target distributions are not observable during training, it becomes challenging for Graph ML models to enhance the generalizability with limited data during training.
Second, data threats stem from encountering unseen data distributions that were not represented during training, potentially impairing the performance of Graph ML models in novel scenarios \cite{li2022learning}. Given that target distributions are not observable during training, it becomes challenging for Graph ML models to enhance generalizability with the limited data available.
%Notably, the structural diversity in graph data makes it difficult for Graph ML models to generalize across different graph distributions without specific strategies to manage this diversity, such as capturing consistent patterns across distributions~\cite{miao2022interpretable}.
%
% Third, evasion attacks also represent a substantial threat to generalizability for Graph ML models. These attacks deliberately manipulate input data during inference, in order to induce model prediction errors and undermine model safety~\cite{feng2019graph,liu2022trustworthy}.
Third, evasion attacks pose a significant threat to the generalizability of Graph ML models. These attacks involve deliberately manipulating input data during inference to induce prediction errors and compromise model safety \cite{feng2019graph, liu2022trustworthy}.
% Furthermore, in graph data, adversarial perturbations are more harmful and harder to detect, as altering only a small number of edges or node features can propagate errors across the graph, due to its structural nature~\cite{liu2022trustworthy}. 
%
% In concrete, these threats pose risks to model generalizability from different perspectives, thus requiring specific safeguard techniques to deal with them.
% Below we introduce safeguard strategies that enhance the generalizability of Graph ML models.
These threats compromise model generalizability from various angles, necessitating tailored safeguard techniques to address them. Below, we introduce strategies that enhance the generalizability of Graph ML models.

\begin{table*}[!t]\centering
\caption{
Categories of methods for Graph ML out-of-distribution generalization.
}\vspace{-0.05in}
\setlength{\tabcolsep}{10.9pt}
\renewcommand{\arraystretch}{1.15}
\scalebox{1}{
  \setlength{\aboverulesep}{0pt}
  \setlength{\belowrulesep}{0pt}
  \begin{tabular}{c|c|ccc|c|c}
    \midrule[1pt]
\multirow{2}{*}{\textbf{Category}}&\multirow{2}{*}{\textbf{Strategy}}&\multicolumn{3}{c|}{\textbf{Task}}& \textbf{Require}&\multirow{2}{*}{\textbf{References}}\\\cline{3-5}
&&Node-level&Graph-level& Size-shift & \textbf{Domain Labels}\\\hline
\multirow{3}{*}{Adversarial}&\multirow{2}{*}{Adversarial}& \redcheck&\rule[0.5ex]{0.6em}{1pt}&\rule[0.5ex]{0.6em}{1pt} & \rule[0.5ex]{0.6em}{1pt} &\cite{xue2021cap,feng2019graph,sadeghi2021distributionally}\\ 
&\multirow{2}{*}{Training}& \rule[0.5ex]{0.6em}{1pt}&\redcheck&\rule[0.5ex]{0.6em}{1pt} & \redcheck&\cite{wu2019domain}\\
&&\redcheck&\redcheck&\rule[0.5ex]{0.6em}{1pt}&\rule[0.5ex]{0.6em}{1pt}&\cite{wu2023adversarial}\\\hline
\multirow{5}{*}{Invariant}& \multirow{3}{*}{Optimization} & \redcheck&\rule[0.5ex]{0.6em}{1pt}&\rule[0.5ex]{0.6em}{1pt} & \redcheck&\cite{wu2022handling,zhu2021shift,zhu2023mario}\\
&& \rule[0.5ex]{0.6em}{1pt}&\redcheck&\rule[0.5ex]{0.6em}{1pt} & \redcheck&\cite{wu2022discovering,li2022learning,miao2022interpretable} \\
&&\redcheck&\redcheck&\rule[0.5ex]{0.6em}{1pt}&\rule[0.5ex]{0.6em}{1pt}&\cite{yang2023individual,yu2023mind} \\\cline{2-7}
&\multirow{2}{*}{Causality-based}& \rule[0.5ex]{0.6em}{1pt}&\redcheck&\rule[0.5ex]{0.6em}{1pt} & \rule[0.5ex]{0.6em}{1pt}&\cite{chen2022learning} \\
&& \rule[0.5ex]{0.6em}{1pt}&\redcheck&\redcheck & \rule[0.5ex]{0.6em}{1pt}&\cite{sui2022causal,bevilacqua2021size}\\
    \midrule[1pt]
  \end{tabular}}
  \vspace{-.1in}
  \label{tab:ood}
\end{table*}

\input{generalizability_model}

\subsection{Data Threat: Distribution Shifts}

% The data threat to generalizability, caused by distribution shift, highlights a fundamental challenge in safe Graph ML: the model's ability to adapt to data that differs significantly from the training distribution~\cite{li2022out,li2022learning}. This type of threat emerges when a model, trained on a specific dataset, is applied to new data that follows a different distribution. 

% To deal with distribution shifts, out-of-distribution (OOD) generalization has become a pivotal research avenue for safe Graph ML, especially in high-stakes domains like medicine, finance, and criminal justice~\cite{miao2022interpretable,yang2023individual,yu2023mind}. 
% %
% Existing OOD generalization techniques include two categories:
The data threat to generalizability, caused by distribution shifts, underscores a core challenge in safe Graph ML: the ability of a model to adapt to data that significantly deviates from the training distribution \cite{li2022out, li2022learning}. This type of threat arises when a model trained on a specific dataset is subsequently applied to new data following a different distribution. To address distribution shifts, out-of-distribution (OOD) generalization has emerged as a critical area of research in Graph ML, particularly in high-stakes fields such as medicine, finance, and criminal justice \cite{miao2022interpretable, yang2023individual, yu2023mind}. Existing techniques for OOD generalization fall into two main categories:
(1) \textit{Adversarial Training} methods aim to enhance the consistency of Graph ML model performance across distributions by intentionally introducing larger distribution shifts during training~\cite{wu2019domain,li2022out}. 
(2) \textit{Invariant Learning} methods utilize the principle of invariant learning, which posits that there is a consistent, invariant component across distributions for each sample. These methods aim to identify and extract this component to facilitate out-of-distribution (OOD) generalization \cite{sui2022causal, yang2023individual, zhu2023mario}. Table~\ref{tab:ood} shows representative works in OOD generation for Graph ML.

\subsubsection{Adversarial Training}

GNN-DRO \cite{sadeghi2021distributionally} employs distributionally robust optimization \cite{rahimian2019distributionally} to address out-of-distribution (OOD) generalization challenges in node classification. Specifically, it minimizes the worst-case loss under the assumption that the data distribution fits within a Wasserstein ball centered around the empirical data distribution.
For graph-level tasks, various adversarial methods aim to stabilize model performance across different domains. For instance, DAGNN \cite{wu2019domain} seeks to learn generalizable graph representations inspired by the DANN architecture \cite{ganin2016domain}. In particular, it achieves this by performing adversarial learning between a domain classifier and an encoder, thus enhancing model generalizability.
However, DAGNN necessitates domain labels for each sample, which may not always be practical \cite{wu2022handling}. To overcome this limitation, GraphAT \cite{feng2019graph} introduces perturbations to specific graphs to create augmented domains. It generates adversarial perturbations for a sample by amplifying the differences between the predictions of the target sample and its neighbors. These perturbations are designed to significantly alter the graph smoothness of the sample, thereby creating new domains for optimization.

\subsubsection{Invariant Learning}

Invariant learning methods focus on exploiting the stable relationships between graphs and labels that persist across different data distributions while disregarding inconsistent or spurious correlations~\cite{ahuja2020invariant}. At its core, invariant learning assumes that each sample is composed of two elements: (1) an invariant component, which maintains consistent feature-label relationships across various distributions, and (2) a variant component, where these relationships might change with shifts in the data distribution. By relying solely on the invariant component for making predictions, models are better equipped to handle significant distribution shifts, thereby enhancing out-of-distribution (OOD) generalization capabilities \cite{muandet2013domain, arjovsky2019invariant, creager2021environment}.

For example, DIR \cite{wu2022discovering} uses a GNN-based generator to identify invariant subgraphs from graphs under interpolated data distributions to isolate stable graph-label relationships across distributions by eliminating the variant part. DIR does not require separating training data based on their distributions, as it utilizes interpolated distributions. In contrast, GIL \cite{li2022learning} leverages the separation information of training data. It identifies invariant subgraphs by clustering variant subgraphs according to how the training data is divided, and it employs an invariant learning module to generate invariant graph representations for classification. Differently, CIGA \cite{chen2022learning} delves into the causality behind invariant subgraphs. By focusing on the direct causal factors of the labels, CIGA enhances OOD generalization by extracting and utilizing these critical, invariant components across different distributions. More recent works propose to exploit invariant information from perspectives like the information bottleneck~\cite{yang2023individual} or contrastive learning~\cite{zhu2023mario}.

\subsection{Attack Threat: Model Evasion} \label{sec:model evasion attacks}
% Model evasion attacks pose a significant challenge to generalizability in safe Graph ML during inference by intentionally perturbing the input or corrupting the utility of model predictions~\cite{jin2019power,feng2019graph}. Therefore, this threat is more severe in critical applications like drug discovery~\cite{lu2020decentralized} or financial analysis~\cite{yang2019ffd}, where incorrect predictions could lead to far-reaching consequences. 
Model evasion attacks significantly challenge the generalizability of safe Graph ML during inference by deliberately perturbing the input or corrupting the utility of model predictions \cite{jin2019power,feng2019graph}. This threat is particularly severe in critical applications such as drug discovery \cite{lu2020decentralized} and financial analysis \cite{yang2019ffd}, where incorrect predictions could have far-reaching consequences.
%
% Below we delineate the two aspects of addressing model evasion attacks on Graph ML: attack mechanisms and defense strategies. We provide a more detailed categorization in Table~\ref{tab:evasion_attack}.
Below, we outline two key aspects of addressing model evasion attacks on Graph ML: attack mechanisms and defense strategies. A more detailed categorization can be found in Table~\ref{tab:evasion_attack}.

\subsubsection{Model Evasion Attacks}

In model evasion attacks, attackers aim to maximize an objective of a trained Graph ML model $f^{*}$.
We formulate the problem of model evasion attacks on graphs as below.
\begin{problem}\label{prob:evasion attack}
Given a graph $\mathcal{G}=(\mathcal{V},\mathcal{E})$, let $f:\mathcal{G}\rightarrow Z$ be a graph machine learning model, where $Z$ denotes the node embedding.
The model evasion attacks on graphs can be formulated as
\begin{equation}
\begin{aligned}
\max_{\mathcal{G}^{\prime}\in\mathcal{F}}\mathcal{L}_{atk}(\mathcal{G}^{\prime},f^{*}),\quad
\st \;&f^{*}=\argmin_f\mathcal{L}(\mathcal{G},f),
\end{aligned}
\end{equation}
where $\mathcal{L}_{atk}$ denotes the attacker's objective function and $\mathcal{F}$ denotes the feasible space of poisoned graphs.
\end{problem}
Below, we introduce attacks on node classification, link prediction, and other tasks on graphs.
% Intuitively, attackers aim to maximize an objective of a normally trained $f^{*}$.
% Below we introduce attacks on node classification, link prediction, and other tasks on graphs.

\begin{table}[!t]
\centering
\caption{Different categories of methods for evasion attacks on Graph ML.}
\vspace{-0.05in}
\setlength{\tabcolsep}{4.9pt}
\renewcommand{\arraystretch}{1.2}
\scalebox{1}{
\setlength{\aboverulesep}{0pt}
\setlength{\belowrulesep}{0pt}
\begin{tabular}{c|c|c}
\toprule[1pt]
 \textbf{Target Task} & \textbf{Setting} & \textbf{References} \\
\midrule
 \multirow{2}{*}{Community Detection} & White-box & \cite{chen2018fast} \\
& Black-box & \cite{chen2017practical,chen2019ga,chen2018fast} \\
\cline{1-3}
\multirow{2}{*}{Network Embedding} & White-box & \cite{chen2018fast} \\ & Black-box & \cite{chen2018fast,dai2018adversarial} \\
\cline{1-3}
\multirow{2}{*}{Node/Graph} & White-box & \cite{wu2019adversarial,zhang2021projective} \\
\multirow{2}{*}{Classification} & Gray-box & \cite{zugner2018adversarial} \\
& Black-box & \cite{chang2020restricted,wan2021adversarial,mu2021hard,ma2020towards} \\
\cline{1-3}
Link Prediction & Gray-box & \cite{lin2020adversarial} \\
\cline{1-3}
Trustworthiness & Black-box & \cite{hussain2022adversarial} \\
\bottomrule[1pt]
\end{tabular}}
\vspace*{-.15in}
\label{tab:evasion_attack}
\end{table}

\vspace{0.02in}
\noindent\textbf{Node Classification.}
%
% {IG-JSMA}~\cite{wu2019adversarial} is a white-box attack that exploits the integrated gradient to estimate the influence of flipping an edge to the attacker's objective and find the edge that increases the attacker's objective the most in the discrete graph structure domain.
% %
% In {Projective Ranking}~\cite{zhang2021projective}, the perturbed edge is chosen based on the mutual information between the attacker's objective and the input graph.
% %
% In a more practical black-box setting, {GF-Attack}~\cite{chang2020restricted} is modeled as corrupting the spectral properties of the graph filter. The attack objective is proved to be transferrable among different types of graph learning models. Such a setting is also followed by other recent works~\cite{wan2021adversarial,mu2021hard,ma2020towards}.
IG-JSMA \cite{wu2019adversarial} is a white-box attack that utilizes integrated gradients to estimate the impact of flipping an edge on the attacker's objective, identifying the edge that maximizes this objective within the discrete graph structure domain.
In Projective Ranking \cite{zhang2021projective}, the perturbed edge is selected based on the mutual information between the attacker's objective and the input graph.
In a more applicable black-box setting, GF-Attack \cite{chang2020restricted} aims to disrupt the spectral properties of the graph filter. The attack's objective has been demonstrated to be transferrable across various types of graph learning models. This approach has been adopted by other recent studies \cite{wan2021adversarial, mu2021hard, ma2020towards, zhang2024sd}.

\vspace{0.02in}
\noindent\textbf{Link Prediction.}
Lin et al. \cite{lin2020adversarial} propose two heuristic gray-box attacks, GGSP and OGSP. GGSP employs a greedy strategy to perturb the edge that results in the largest decrease in link prediction performance, while OGSP selects edges based on an analysis of common neighbors.
Additionally, preliminary studies have highlighted vulnerabilities related to fairness \cite{zeng2021fair, mehrabi2021exacerbating}, interpretation \cite{ghorbani2019interpretation, jain2019attention}, and privacy \cite{tramer2022truth} in tabular, image, or text data. However, the unique interdependencies between graph components and the specific mechanisms of Graph ML models mean that attacks on the trustworthiness of graph data remain relatively unexplored.
FA-GNN \cite{hussain2022adversarial} specifically targets the algorithmic fairness of graph learning models by injecting adversarial links. This method categorizes nodes into four groups based on the label and sensitive attribute information and introduces random cross-group links, thereby highlighting the susceptibility of fairness in Graph ML models.

\noindent\textbf{Other Tasks.}
We introduce evasion attacks on additional tasks such as community detection \cite{chen2017practical, chen2019ga} and network embedding \cite{chen2018fast, dai2018adversarial}.
In the context of community detection, Chen et al. \cite{chen2017practical} present the first unsupervised evasion attack with two distinct strategies. The first strategy involves injecting new nodes that mimic target nodes to mislead the clustering process. The second strategy randomly removes nodes and edges from a graph extended from the original nodes.
Q-Attack \cite{chen2019ga} employs heuristic algorithms to rewire edges in the original graph in a black-box setting, optimizing the attack to disrupt community structures.
For network embedding, FGA \cite{chen2018fast} alters graph connections by selecting the largest gradient component of the attacker's objective, using cross-entropy loss on target nodes as the loss function. FGA is capable of attacking both community detection and network embedding models. Beyond using gradient information, reinforcement learning is utilized to develop an effective attack strategy. RL-S2V \cite{dai2018adversarial} is a black-box model evasion attack that uses the Markov Decision Process to model sequences of graph perturbations, with Q-learning employed to optimize the attack's reward function.

\subsubsection{Model Evasion Defense}
In this part, we introduce various strategies proposed for defense against model evasion attacks on Graph ML.

\vspace{0.02in}
\noindent \textbf{Adversarial Training.} 
% The rationale of adversarial training is to mimic the attacker by maximizing the task-oriented loss during training~\cite{sun2022adversarial}. Then the Graph ML model is empirically more robust against model evasion attacks during inference. 
The rationale behind adversarial training is to simulate an attacker by maximizing the task-oriented loss during the training process \cite{sun2022adversarial}. As a result, this approach empirically enhances the robustness of Graph ML models against model evasion attacks during inference. The corresponding objective is given as
\begin{align}
    \theta^*=\argmin_{\theta}\  \max_{\mathcal{G}} \mathcal{L}(f_{\theta}(\mathcal{G}), \mathcal{Y}).
\end{align}
Here $f$ is a Graph ML model parameterized by $\theta$; $\mathcal{G}$ is the graph used for training; $\mathcal{L}$ denotes the loss function; $\mathcal{Y}$ represents the ground truth. 
Numerous existing studies have proposed effective defense strategies based on the rationale of adversarial training. For instance, in \cite{dai2018adversarial}, the authors suggest randomly dropping edges during training, a method that introduces topological noise and may improve the objective function. This straightforward strategy has been empirically shown to enhance the robustness of GNNs in both graph and node classification tasks. Additionally, topological perturbations can be further optimized using projected gradient descent \cite{xu2019topology}, which has been demonstrated to fortify GNNs without compromising model utility. Other studies have similarly employed perturbations to either graph topology \cite{dai2019adversarial} or node attributes \cite{jin2019latent,wang2019graphdefense} to improve defense mechanisms.

\vspace{0.02in}
\noindent \textbf{Robustness Certification.}
%Although adversarial training can achieve defense empirically, they are not able to provide certification. Therefore, once a certain methodology is proposed, they may be subsequently defeated by the newly proposed attacking methods
% While adversarial training achieves defense empirically, it lacks formal robustness certification, making the model vulnerable to newly proposed attack methods~\cite{jin2021adversarial}. In light of this, multiple recent works propose to achieve certification over the robustness of Graph ML models. For example, Bojchevski et al. ~\cite{bojchevski2019certifiable} proposed a certified defense strategy for GNNs against structural perturbations, while Zugner et al. ~\cite{zugner2019certifiable} proposed to achieve certified defense against node feature perturbations. Further advancements have enabled certified defenses against both types of perturbations~\cite{dong2023elegant}. 
While adversarial training has shown empirical success in defense, it lacks formal robustness certification, leaving the model potentially vulnerable to newly developed attack methods \cite{jin2021adversarial}. In response to this limitation, several recent studies have focused on achieving robustness certification for Graph ML models. For instance, Bojchevski et al. \cite{bojchevski2019certifiable} introduced a certified defense strategy for GNNs against structural perturbations, and Zugner et al. \cite{zugner2019certifiable} developed methods for certified defense against node feature perturbations. Subsequent advancements have led to certified defenses capable of addressing both types of perturbations \cite{dong2023elegant}.
%Notably, recent research has expanded certified defenses to tasks like community detection~\cite{jia2020certified}.

%It is worth noting that most works above only focus on the node classification task, where such certification is achieved over a given specific node. Recent studies have extended certified defense to other tasks such as community detection~\cite{jia2020certified}.

\vspace{0.02in}
\noindent \textbf{Data \& Model Refinement.}
% Defense against model evasion attacks can be enhanced by altering input graph data or the graph learning model itself. For data manipulation, techniques include preprocessing~\cite{ioannidis2019graphsac} and augmentation~\cite{fox2019robust} to safeguard against attacks. On the model side, replacing the traditional GCN's graph Laplacian with a graph powering operator~\cite{jin2019power} has been shown to enhance GNN robustness against evasion attacks.
Defense against model evasion attacks can be strengthened either by modifying the input graph data or by adapting the Graph ML model itself. For data manipulation, techniques such as preprocessing \cite{ioannidis2019graphsac} and augmentation \cite{jin2021adversarial} are employed to protect against attacks. On the model front, replacing the traditional graph Laplacian in GCNs with a graph powering operator \cite{jin2019power} has proven to enhance the robustness of GNNs against evasion attacks. %augmentation \cite{fox2019robust}

\section{Confidentiality in Graph ML}
% Confidentiality in Graph ML refers to the protection of sensitive information encoded in the data, the model predictions, and the model itself~\cite{olatunji2023releasing}. This protection is vital for both user privacy and legal compliance~\cite{xu2023machine}. For example, in the use of Graph ML models for user categorization on social networks, it is essential to prevent the inadvertent leakage of user information during both training and inference~\cite{chen2020vertically}. 
% Thus, confidentiality presents an important aspect of Graph ML safety.
Confidentiality in Graph ML encompasses the protection of sensitive information within the data, the model predictions, and the model itself \cite{olatunji2023releasing}. This protection is crucial for ensuring user privacy and meeting legal compliance requirements \cite{xu2023machine}. For instance, when employing Graph ML models for user categorization on social networks, it is imperative to prevent the inadvertent leakage of user information during both training and inference stages \cite{chen2020vertically}.

\begin{figure}\center
\includegraphics[width=0.4\textwidth] {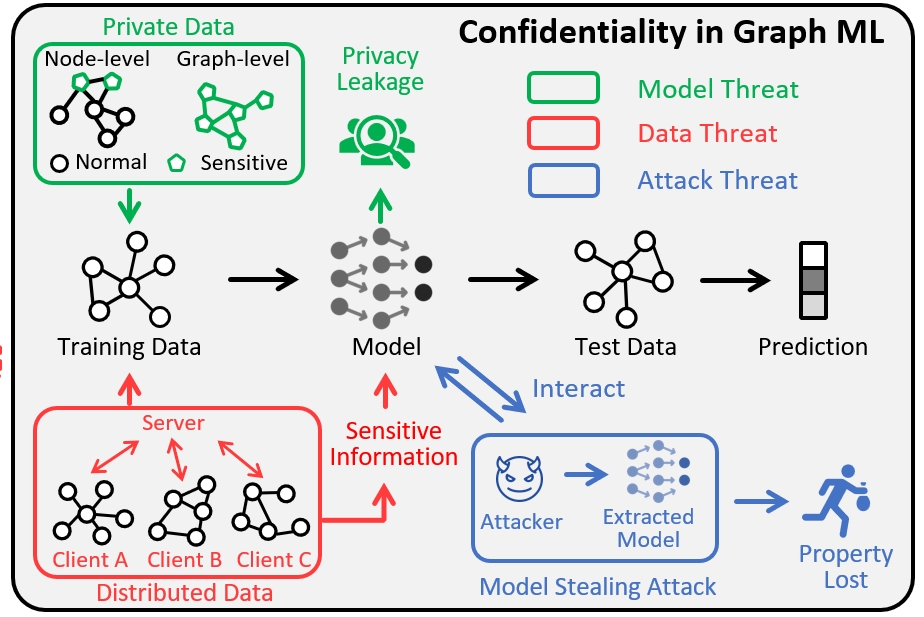}
\vspace*{-.05in}
\caption{Safety threats regarding confidentiality in Graph ML.}
\vspace*{-.15in}
\label{fig:confidentiality}
\end{figure}

% Here we introduce three types of threats against confidentiality, as shown in Fig.~\ref{fig:confidentiality}. First, the model threat arises from the inherent model design that could potentially leak sensitive information via the model's predictions or architecture, thereby causing privacy concerns~\cite{wu2022linkteller}. This is particularly concerning on graphs, 
% where the message-passing mechanism can reveal sensitive information from neighboring nodes~\cite{chen2020vertically,daigavane2021node,sajadmanesh2021locally}.
Here, we introduce three types of threats to confidentiality as depicted in Fig.~\ref{fig:confidentiality}. First, the model threat arises from inherent design flaws in the model that could potentially leak sensitive information through its predictions or architecture, thus raising privacy concerns \cite{wu2022linkteller}. This issue is particularly acute in graph-based models, where the message-passing mechanism may inadvertently expose sensitive information from neighboring nodes \cite{chen2020vertically, daigavane2021node, sajadmanesh2021locally}.
Second, the data threat emerges when training data is distributed across multiple sources, potentially leading to the unintentional exposure of sensitive information \cite{huang2023federated}. Given the structural nature of graph data, ensuring data confidentiality across each source presents a significant challenge \cite{fedpub, lei2023federated}. Third, the attack threat directly endangers model confidentiality through unauthorized cloning of the model or its functionality \cite{xu2023watermarking}. This not only breaches privacy but also violates intellectual property rights, as the replicated model could be used or sold without permission \cite{defazio2019adversarial, wu2022model}. Below, we will explore protective strategies designed to mitigate these confidentiality threats in Graph ML.

%Below we thoroughly investigate the safeguard techniques designed to mitigate the impacts of these threats for confidentiality in Graph ML.

% This aspect of safety is of paramount importance, especially when Graph ML models process data that may include personal identifiers, financial records, health information, or any other type of data that should remain private~\cite{dai2022comprehensive}. In these cases, maintaining confidentiality is not only a matter of user privacy but also of legal compliance~\cite{xu2023machine}. 

\subsection{Model Threat: Privacy}
% The model threat against confidentiality stems from model designs that could potentially expose private information through the model architecture~\cite{mei2019sgnn} and learning mechanisms~\cite{cong2022grapheditor}. For example, the message-passing mechanism in GNNs
% may unintentionally disseminate a node's sensitive information to neighboring nodes,
% potentially leading to privacy leakgage~\cite{dai2022comprehensive,zhu2020mixedad}.  
% Here we focus on two prevalent groups of privacy-preserving technologies: \textit{Differential Privacy} and \textit{Graph Unlearning}. A detailed categorization of these methods is provided in Table~\ref{tab:privacy_subtable}.
%the sensitive information of a specific node could be unintentionally propagated to its neighbors, 

The model threat to confidentiality arises from designs that could inadvertently expose private information through the architecture \cite{mei2019sgnn} and learning mechanisms of the model \cite{cong2022grapheditor}. For instance, the message-passing mechanism in GNNs might unintentionally share a node’s sensitive information with neighboring nodes, leading to potential privacy breaches \cite{dai2022comprehensive, zhu2020mixedad}. In this discussion, we focus on two prominent categories of privacy-preserving technologies: \textit{Differential Privacy} and \textit{Graph Unlearning}. A detailed categorization of these methods is provided in Table~\ref{tab:privacy_subtable}.

% Specifically, differential privacy (DP) offers a robust framework for quantifying and mitigating the risk of privacy breaches. It provides a mathematical guarantee that the inclusion or exclusion of a single data point will not significantly impact the outcome of a computation, thereby preserving individual privacy within a dataset~\cite{dwork2006differential}. 
% Graph unlearning, on the other hand, addresses the challenge of undoing the impact of previously learned information from Graph ML models~\cite{cao2015towards,baumhauer2022machine,nguyen2022survey}. In scenarios where data points need to be revoked or forgotten, graph unlearning allows models to adapt without extensive retraining. 

\begin{table*}[!t]
\centering
\caption{Different categories of privacy-preserving Graph ML methods.}
\vspace{-0.05in}
\setlength{\tabcolsep}{10.9pt}
\renewcommand{\arraystretch}{1.2}
\scalebox{1}{
  \setlength{\aboverulesep}{0pt}
  \setlength{\belowrulesep}{0pt}
  \begin{tabular}{c|c|c|c|c}
    \midrule[1pt]    
    \textbf{Category} & \textbf{Unlearning Strategy} & \textbf{Graph Type} & \textbf{Task} & \textbf{References} \\\hline
    \multirow{4}{*}{Graph Unlearning} & Exact Unlearning & Undirected & Node-level & \cite{chen2022graph,cong2022grapheditor} \\\cline{2-5}
    & \multirow{3}{*}{Approximate Unlearning} &\multirow{3}{*}{Undirected} & Node-level & \cite{chien2023efficient,cheng2023gnndelete,wu2023gif} \\
    &&& Edge-level & \cite{wu2023certified,chien2023efficient} \\
    &&& Graph-level & \cite{pan2023unlearning} \\\hline
    \multirow{3}{*}{Differential Privacy} & \multirow{3}{*}{\rule[0.5ex]{0.6em}{1pt}} & Undirected & \multirow{2}{*}{Node-level} & \cite{mei2019sgnn,sajadmanesh2021locally,olatunji2023releasing,chen2020vertically,sajadmanesh2023gap,xu2018dpne} \\\cline{3-3}
    && Directed & & \cite{chien2023differentially,daigavane2021node} \\\cline{3-5}
    && Undirected & Edge-level & \cite{wu2022linkteller,sajadmanesh2023gap,zhang2019graph,xu2018dpne} \\  
    \midrule[1pt]
  \end{tabular}}
  \vspace*{-.1in}
  \label{tab:privacy_subtable}
\end{table*}

\subsubsection{Differential Privacy}
% Differential privacy (DP) is a concept designed to safeguard the privacy encoded in datasets by ensuring that the presence or absence of a single sample does not substantially influence the outcomes of the model~\cite{chien2023differentially, wu2022survey}. 
Differential Privacy (DP) is a framework designed to protect the privacy of data within datasets by ensuring that the inclusion or exclusion of any single individual's data does not significantly affect the output of the model \cite{chien2023differentially, wu2022survey}.
Generally, a randomized algorithm $\mathcal{M}$ is $(\epsilon,\delta)$-DP if for any sample $\mathbf{z}$ in the training dataset $\mathcal{D}$ and for all $\mathcal{S} \subseteq Range(\mathcal{M})$, it holds that $Pr(\mathcal{M}(\mathcal{D})\in \mathcal{S}) \leq e^{\epsilon}Pr(\mathcal{M}(\mathcal{D}\setminus \mathbf{z})\in \mathcal{S})+\delta$.
Differential Privacy (DP) enhances privacy by adding controlled noise to model updates. For example, DP-SGD \cite{abadi2016deep} involves clipping gradient norms and introducing noise to the gradients. Traditional DP methods, however, are less suited to graph data due to the structural intricacies of graphs. Addressing this challenge, DPNE \cite{xu2018dpne} and PPGD \cite{zhang2019graph} apply DP in learning graph representations. More recent methods have focused on integrating DP with Graph Neural Networks (GNNs). For instance, DP-GNN \cite{daigavane2021node} utilizes a variation of DP-SGD \cite{bassily2014private} and a novel subgraph sampling technique to safeguard privacy in single-layer GNNs. For multi-layer GNNs, LPGNN \cite{sajadmanesh2021locally} employs a Local Differential Privacy (LDP) framework that perturbs node features and includes a denoising layer to maintain accuracy. Conversely, PrivGnn \cite{olatunji2023releasing} focuses on global DP for the entire graph. However, these methods may not be suitable for scenarios where edge privacy is a concern. To address this, LinkTeller \cite{wu2022linkteller} uses edge-level DP algorithms to infer private edge information. To ensure both edge-level and node-level DP, GAP \cite{sajadmanesh2023gap} integrates stochastic noise into the GNN’s aggregation function, aiming to statistically obscure the presence of individual edges or nodes and their connections.
Additionally, there have been efforts to integrate Differential Privacy (DP) into federated learning. SGNN \cite{mei2019sgnn} and VFGNN \cite{chen2020vertically} each take a novel approach by partitioning the entire graph and the computation graph of a GNN across multiple data holders, respectively.

% , leverage a trusted server to aggregate information from various parties and facilitate holistic training.

\subsubsection{Graph Unlearning}

% To preserve privacy regarding sensitive data, another intuitive solution is to remove it from training data~\cite{nguyen2022survey}. However, the connection between a trained model's parameters and the specific training data is not always clear-cut. This makes effectively removing information about specific data from a Graph ML model challenging. To tackle this, 

% To remove certain knowledge encoded in the model, a novel approach called machine unlearning is required, which aims to remove the influence of specific training data without retraining from scratch~\cite{bourtoule2021machine}. Machine unlearning is particularly relevant for scenarios where individuals may request their data to be deleted~\cite{baumhauer2022machine}.
% Here we primarily focus on machine unlearning algorithms tailored for graph data, referred to as graph unlearning~\cite{chen2022graph}, which can be further categorized into exact graph unlearning and approximate graph unlearning.
To eliminate specific knowledge encoded in a model, an approach known as machine unlearning is employed. This method seeks to eradicate the influence of particular training data without the need to retrain the model from scratch \cite{bourtoule2021machine}. Machine unlearning is especially pertinent in situations where individuals request the deletion of their data \cite{baumhauer2022machine}. In this discussion, we focus primarily on machine unlearning algorithms designed for graph data, termed graph unlearning \cite{chen2022graph}. This can be further divided into two categories: \emph{exact graph unlearning} and \emph{approximate graph unlearning}.

\vspace{0.02in}
\noindent\textbf{Exact Graph Unlearning.}
% With a randomized learning algorithm $\mathcal{M}$, we denote the models trained on the original graph dataset $\mathcal{D}$ and the remaining data as $\mathcal{M}(\mathcal{D})$ and $\mathcal{M}(\mathcal{D}\setminus \mathcal{D}_f)$, respectively, where $\mathcal{D}_f$ represents the forgotten dataset. A randomized unlearning mechanism $\mathnormal{U}$ is an exact unlearning process, if for all $\mathcal{S} \subseteq Range(\mathcal{M})$, it holds that $Pr(\mathnormal{U}(\mathcal{D}, \mathcal{D}_f, \mathcal{M}(\mathcal{D}))\in \mathcal{S}) = Pr(\mathcal{M}(\mathcal{D}\setminus \mathcal{D}_f)\in \mathcal{S})$.
Using a randomized learning algorithm $\mathcal{M}$, we denote the models trained on the original graph dataset $\mathcal{D}$ and on the dataset after removing specific data as $\mathcal{M}(\mathcal{D})$ and $\mathcal{M}(\mathcal{D}\setminus \mathcal{D}_f)$, respectively. Here, $\mathcal{D}_f$ represents the dataset to be forgotten. A randomized unlearning mechanism $\mathnormal{U}$ is considered an exact unlearning process if, for all subsets $\mathcal{S}$ within the range of $\mathcal{M}$, the equality $Pr(\mathnormal{U}(\mathcal{D}, \mathcal{D}_f, \mathcal{M}(\mathcal{D}))\in \mathcal{S}) = Pr(\mathcal{M}(\mathcal{D}\setminus \mathcal{D}_f)\in \mathcal{S})$ holds.
% \[Pr(\mathnormal{U}(\mathcal{D}, \mathcal{D}_f, \mathcal{M}(\mathcal{D}))\in \mathcal{S}) = Pr(\mathcal{M}(\mathcal{D}\setminus \mathcal{D}_f)\in \mathcal{S}).\]
% {GraphEraser}~\cite{chen2022graph} is one of the first graph unlearning framework, which partitions the training graph data into subgraphs, and aggregates models trained on different subgraphs for inference. When the forgotten data is revoked, GraphEraser only retrains the model associated with subgraphs in the deleted data. {GraphEditor}~\cite{cong2022grapheditor} proposes an exact unlearning mechanism for linear GNNs without extensive retraining. It transforms the training of GNN into a problem with a closed-form solution, and removes the impact of certain nodes by directly operating model parameters. 
GraphEraser \cite{chen2022graph} is one of the first frameworks in graph unlearning. It segments the training graph data into subgraphs and aggregates models trained on these subgraphs for inference. When data needs to be forgotten, GraphEraser re-trains only the models associated with subgraphs containing the deleted data. GraphEditor \cite{cong2022grapheditor}, on the other hand, proposes an exact unlearning mechanism for linear GNNs that avoids extensive retraining. It reframes the training of GNNs into a problem with a closed-form solution, allowing for the direct modification of model parameters to negate the influence of specific nodes.

\vspace{0.02in}
\noindent\textbf{Approximate Graph Unlearning.}
% Although the exact unlearning on graphs effectively removes the influence of target graph entities, a large time complexity makes it difficult to tackle large-scale applications with frequent removal requests~\cite{chien2023efficient}.
% Different from exact unlearning, approximate graph unlearning aims to update the original model with an approximation of the retrained model, instead of retraining the original model from scratch~\cite{said2023survey，chien2023efficient}.
% As a result, approximate graph unlearning cannot completely remove the information of the target samples as exact graph unlearning does, but is more efficient for large-scale graphs.
% To measure the approximation error of the retrained model, two parameters, $\varepsilon$ and $\delta$, are used to bound the distance between the retrained model and the unlearned model in the probability space, i.e., for all $\mathcal{S} \subseteq Range(\mathcal{M})$, 
"Unlike exact unlearning, approximate graph unlearning seeks to update the original model with an approximation of the retrained model, rather than retraining from scratch \cite{said2023survey, chien2023efficient}. Consequently, while approximate graph unlearning may not completely eliminate the information of the target samples as exact unlearning does, it offers a more efficient solution for large-scale graphs. To assess the approximation error of the retrained model, two parameters, $\varepsilon$ and $\delta$, are used to quantify the proximity between the retrained model and the unlearned model within the probability space, for all $\mathcal{S} \subseteq Range(\mathcal{M})$
\begin{equation}
\begin{aligned}
Pr(\mathnormal{U}(\mathcal{D}, \mathcal{D}_f, \mathcal{M}(\mathcal{D}))\in \mathcal{S})\leq e^{\varepsilon}Pr(\mathcal{M}(\mathcal{D}\setminus \mathcal{D}_f)\in \mathcal{S})+\delta, \\
Pr(\mathcal{M}(\mathcal{D}\setminus \mathcal{D}_f)\in \mathcal{S})\leq e^{\varepsilon}Pr(\mathnormal{U}(\mathcal{D}, \mathcal{D}_f, \mathcal{M}(\mathcal{D}))\in \mathcal{S})+\delta. \\
\end{aligned}
\end{equation}
% {SGC-unlearn}~\cite{chien2023efficient} is one of the first approximate graph unlearning methods.
% %for Generalized PageRank-based GNNs Simple Graph Convolutions (SGC)~\cite{wu2019simplifying}. %and APPNP~\cite{gasteiger2019predict}.
% In particular, it divides approximate graph unlearning into three tasks: node feature unlearning, edge unlearning, and node unlearning. It further derives theoretical guarantees for each task by finding an upper bound for the influence of removing a node feature, an edge, and a node with all associated edges.
% {GST-unlearn}~\cite{pan2023unlearning} 
% %proposes the first approximate unlearning method for Graph Scattering Transform (GST) models. Specifically, it 
% proposes a nonlinear learning pipeline 
% %%%using a Graph Scattering Transform model~\cite{gama2019stability} to extract graph embeddings and a linear classifier for the downstream graph classification task. 
% and leverages an approximate unlearning algorithm based on ~\cite{chien2023efficient} for the proposed pipeline with a theoretical guarantee.
% %
% {GIF}~\cite{wu2023gif} adapts the traditional influence function to the graph domain and proposes the Graph Influence Function to estimate the retrained model efficiently.
%%%(for all of the node feature, edge, and node unlearning). Moreover, theoretical guarantee on approximate graph unlearning is achieved assuming the loss function to be convex. 
Approximate graph unlearning can be categorized into three specific tasks: node feature unlearning, edge unlearning, and node unlearning, each with theoretical guarantees. These guarantees involve calculating an upper bound for the influence of removing a node feature, an edge, and a node along with all its associated edges. GST-unlearn \cite{pan2023unlearning} introduces a nonlinear learning pipeline and incorporates an approximate unlearning algorithm based on \cite{chien2023efficient}, providing theoretical assurances for its effectiveness. GIF \cite{wu2023gif} adapts the traditional influence function to the graph domain, introducing the Graph Influence Function to efficiently estimate the retrained model. 
Other recent works also conducted explorations on utilizing influence functions~\cite{wu2023certified} or layer-wise analysis of GNNs~\cite{cheng2023gnndelete} for graph unlearning.

% {CEU}~\cite{wu2023certified} proposes an approximate edge unlearning method for general graph neural networks. Specifically, the authors followed a similar pipeline as~\cite{chien2023efficient}, but enabling a wider scope for non-linear GNNs. 
% {GNNDelete}~\cite{cheng2023gnndelete} reformulates the problem of approximate graph unlearning as two properties, deletes edge consistency and neighborhood influence, and proposes a novel graph unlearning method learning a layer-wise deletion operator to revise a trained GNN model based on the two properties.

\begin{table*}[!t]
\setlength{\tabcolsep}{10.3pt}
\centering
\caption{The category of federated learning methods.}
\vspace{-0.05in}
\renewcommand{\arraystretch}{1.2}
\scalebox{1}{
  \setlength{\aboverulesep}{0pt}
  \setlength{\belowrulesep}{0pt}
\begin{tabular}{c|c|c|c}
    \midrule[1pt]    
    \textbf{Category}                       & \textbf{Task}                     & \textbf{Data on Each Client}  & \textbf{References} \\\hline
    \multirow{3}{*}{Data Heterogeneity}     & Graph Classification/Regression   & Multiple graphs               & \cite{xie2021gcfl,fedstar,zhu2022flitplus, lou2021stfl} \\\cline{2-4}
                                            & \multirow{2}{*}{Node Classification}               & A subgraph                    & \cite{fedsage, fedpub, lei2023federated, wang2022graphfl, yao2022fedgcn, zhang2021ppsgcn} \\\cline{3-3}\cline{4-4}
                                            &                                  & A graph                       & \cite{fedlit, huang2023federated} \\\hline
    \multirow{2}{*}{Overlapping Node Alignment}  & Knowledge Graph Completion        & A knowledge graph             & \cite{chen2021fede,zhang2022fedr,peng2021fkge} \\\cline{2-4}
                                            & Rating Prediction                 & A user-item graph             & \cite{liu2021fesog,wu2022fedpergnn} \\\midrule[1pt]
\end{tabular}}
\vspace*{-.1in}
  \label{tab:federated}
\end{table*}

% \subsection{Federated Learning (Dispensability)}
\subsection{Data Threat: Distributed Data}
%We now introduce the unique confidentiality problems posed by distributed data during the training of Graph ML models~\cite{fedsage}. 
% In many real-world applications, due to privacy concerns, graph data is generally decentralized, spanning across multiple clients~\cite{fu2022fgml}. Such a distributed scenario presents significant risks of unauthorized data exposures via the message-passing mechanism during training~\cite{fedsage}. Recently, Federated Graph Learning (FGL) has emerged as a prominent approach for learning from distributed graph data while safeguarding privacy~\cite{zhu2022flitplus,yao2022fedgcn}. Below we first outline the general setting for FGL.
In many real-world applications, privacy concerns necessitate decentralized graph data, often spanning multiple clients \cite{fu2022fgml}. This distributed arrangement poses significant risks of unauthorized data exposure through the message-passing mechanism during training \cite{fedsage}. Recently, Federated Graph Learning (FGL) has emerged as a key approach to learning from distributed graph data while maintaining privacy \cite{zhu2022flitplus, yao2022fedgcn}. Below, we outline the general framework for FGL.

% Federated Learning (FL) addresses these issues and has proven successful in various applications~\cite{aledhari2020federated}. However, well-known FL algorithms like FedAvg~\cite{mcmahan2017communication} are evaluated on tasks within the purview of image and text data. When employed to optimize GNNs, their efficacy remains unexplored~\cite{wang2022federatedscope}.

% Assume that there are $K$ clients, and each client $k$ has its private graph dataset $\mathcal{D}_k$, with $N_k$ the number of samples in client $k$ and $N$ the total number of training samples. The optimization objective of FGL problems can be written as
% \begin{equation}  \label{fl}
%     \min\limits_{(\theta_1,\theta_2,\cdots,\theta_K)}\sum_{k=1}^{K} \frac{N_k}{N}\mathcal{L}_k(\mathcal{D}_k;\theta_k),
% \end{equation}
% where $\mathcal{L}_k$ and $\theta_k$ are local objective function and model parameter in client $k$, respectively. The standard FL aims to learn a global model such that $\overline{\theta}=\theta_1=\theta_2=\cdots=\theta_K$. In the realm of FGL, the algorithmic challenges can be categorized into two perspectives: managing data heterogeneity and ensuring entity alignment across clients. In Table~\ref{tab:federated}, we summarise representative existing studies of FGL.
Assume there are $K$ clients, each possessing a private graph dataset $\mathcal{D}_k$. Let $N_k$ represent the number of samples in client $k$ and $N$ the total number of training samples across all clients. The optimization objective for Federated Graph Learning (FGL) can be expressed as:
\begin{equation}
\min\limits_{(\theta_1,\theta_2,\cdots,\theta_K)}\sum_{k=1}^{K} \frac{N_k}{N}\mathcal{L}_k(\mathcal{D}_k;\theta_k),
\end{equation}
where $\mathcal{L}_k$ and $\theta_k$ denote the local objective function and model parameters for client $k$, respectively. Standard Federated Learning (FL) aims to develop a global model where $\overline{\theta}=\theta_1=\theta_2=\cdots=\theta_K$. Within the scope of FGL, challenges mainly revolve around managing data heterogeneity and ensuring node alignment across different clients. Representative studies in FGL are summarized in Table~\ref{tab:federated}.

\subsubsection{Data Heterogeneity}
% Previous endeavors for handling data heterogeneity can be categorized into two groups, namely subgraph-level and graph-level methods, based on the types of data held by clients and the specific target tasks they aim to achieve. 
Previous efforts to address data heterogeneity in Federated Graph Learning can be divided into two categories: subgraph-level and graph-level methods. This classification is based on the types of data possessed by the clients and the specific target tasks they aim to accomplish.

\vspace{0.02in}
\noindent\textbf{Subgraph-Level Methods.}
%FedCog can attain equivalent node representations as aggregation on the entire graph but in a privacy-preserving manner. 
In subgraph-level Federated Graph Learning (FGL), each client holds a subgraph of the entire graph. This framework is tailored for tasks such as neighbor generation \cite{fedsage} and community discovery \cite{fedpub}. FedSage \cite{fedsage}, for example, addresses the challenge of missing links in supervised learning tasks. It involves a collaborative effort to train neighbor generators, $\theta^{eg}_k$, for each client. These generators are then used by clients to reconstruct their subgraphs, which are subsequently used to train a GraphSage model \cite{hamilton2017inductive}. However, FedSage can be sensitive to the initial settings of the generative model and prone to node mismatches. To overcome these issues, FedCog \cite{lei2023federated} decouples local subgraph computations and differentiates propagation across inter and intra-edges.
Recent developments in subgraph-level FGL have also delved into semi-supervised optimization \cite{wang2022graphfl, yao2022fedgcn} and structural heterogeneity \cite{fedlit, fedpub}, expanding the applications and capabilities of FGL.

\vspace{0.02in}
\noindent\textbf{Graph-Level Methods.}
These methods operate under the assumption that clients each possess distinct graphs and generally seek to leverage shared properties across these graphs \cite{leskovec2005realistics, qiu2020gcc, fedstar, xie2021gcfl}. For example, GCFL \cite{xie2021gcfl} dynamically groups clients into clusters based on gradients in GNNs, allowing clients within the same cluster to share and update a common cluster model rather than a single global model. Unlike approaches that share the entire GNN among clients, FedStar \cite{fedstar} focuses on the structural properties common across different domains. It divides the model into a structure encoder, $\theta^{s}$, and a feature encoder, $\theta^{f}$, with $\theta = (\theta^{s},\theta^{f})$. FedStar exclusively shares $\theta^{s}$ among clients and redefines the optimization problem as follows:
\begin{equation}\min\limits_{(\overline{\theta}^{s},\theta^{f}_1,\theta^{f}_2,\cdots,\theta^{f}_K)}\sum_{k=1}^{K} \frac{N_k}{N}\mathcal{L}_k(\mathcal{D}_k;(\overline{\theta}^{s},\theta^{f}_k)).
\end{equation}
There are also domain-specific methods, e.g., those for molecular learning~\cite{zhu2022flitplus} and spatial-temporal predictions~\cite{lou2021stfl}.

\subsubsection{Overlapping Node Alignment}
% In federated settings, the same entity can appear across multiple clients, referred to as overlapping entities. In generic FL, studies have addressed this by aligning the embeddings of overlapping entities~\cite{chen2021fede, peng2021fkge, zhang2022fedr}. Generally, the techniques in FGL for overlapping entities alignment fall into two categories: knowledge graph (KG)-based alignment and user-item graph-based alignment.

In federated settings, a node in a client may also appear in other clients, known as an overlapping node. Taking a recommendation system as an example, a user may possess items that it has interactions with. In the meantime, these items can also interact with other users. In this case, the embedding of a specific item should be consistent across users.
Prevalent FGL techniques attempt to achieve this by aligning the embeddings of these overlapping nodes across clients \cite{chen2021fede, peng2021fkge, zhang2022fedr}. A common approach involves a central server that maintains a global node embedding matrix \cite{chen2021fede}. The server updates the matrix using local embedding matrices received from each client and then broadcasts these global embeddings back to the clients for further local training.
In such scenarios, it is crucial to preserve edge information on the client side. Simply aligning the embeddings of overlapping nodes on the server can lead to privacy risks since the server could potentially infer edge information by observing nodes associated with non-zero-gradient embeddings, especially in recommendation systems. This occurs because an item embedding is only updated when there is a link to the user \cite{wu2022fedpergnn}.
To address this issue, a widely adopted technique is pseudo-interacted item sampling \cite{liu2021fesog}. Typically, before sending gradients to the central server, each client samples some items with which they have not interacted (i.e., pseudo-interacted items). Subsequently, users generate embedding gradients for these sampled items (e.g., using a Gaussian distribution) and merge them with the gradients of the real interactions.

\subsection{Attack Threat: Model Stealing}

Optimized Graph ML models represent significant intellectual property (IP) for model owners, given the substantial investments required in constructing training datasets, tuning hyperparameters, and optimizing learnable parameters \cite{frasca2020sign, waheed2023grove}. However, these models are vulnerable to theft, as malicious entities might replicate the model or misappropriate its functionalities \cite{orekondy2019knockoff, tramer2016stealing}. Consequently, establishing verifiable ownership of Graph ML models is a pressing necessity, drawing considerable research interest. In this section, we explore (1) attack methods for stealing Graph ML models, and (2) defense strategies to prevent such theft.

% Knockoff nets: Stealing functionality of black-box models

% Practical black-box attacks against machine learning.

% Stealing machine learning models via pre- diction {APIs}.

\subsubsection{Model Stealing Attack}

% Given a target model originally trained by the model owner,  model stealing attacks aim to replicate its functionalities by training a surrogate model through interactions with the target model~\cite{defazio2019adversarial}. Such a surrogate model can often steal certain functionalities of the target model (e.g., achieving similar node classification accuracy or similar predictive behavior of the target model), such that the attacker can obtain a model similar to the target model with minimal  efforts~\cite{defazio2019adversarial,wu2022model,shen2022model}.
% %
% This section focuses primarily on node classification, the most prevalent task targeted by these attacks. 
Model stealing attacks are designed to replicate the functionalities of a target model, originally trained by the model owner, by training a surrogate model through interactions with the target model \cite{defazio2019adversarial}. Often, such a surrogate model can appropriate certain functionalities of the target model, such as achieving comparable node classification accuracy or mimicking the predictive behavior of the target model, allowing the attacker to obtain a similar model with minimal effort \cite{defazio2019adversarial, wu2022model, shen2022model}.
This section will primarily focus on node classification, which is the most common task targeted by these attacks.
%The most representative and widely studied task under such attack is node classification. Therefore, we mainly focus on the node classification task in this section.
%
% defazio2019adversarial
% wu2022model
% shen2022model
We first present a formal problem definition.
\begin{problem}\label{stealing_attack}
Given certain knowledge regarding the graph data $\mathcal{G}$, the goal of the attacker is to train a surrogate model $f_{\text{G}}^*$ to match certain characteristics (e.g., similar prediction accuracy or similar predictive behaviors) of a target model $f_{\text{G}}$ by collecting certain query results from $f_{\text{G}}$.
% training graph data $\mathcal{G}$, model stealing defense for graph machine learning aims to achieve an optimized original model $f_{\text{G}}$ trained on $\mathcal{G}$, such that for any model $f_{\text{G}}^*$ trained relying on $f_{\text{G}}$ and $\tilde{f}_{\text{G}}^*$ trained without relying on $f_{\text{G}}$, we are able to find a function $h$ where $h(f_{\text{G}}^*) = 1$ and $h(\tilde{f}_{\text{G}}^*) = 0$.
\end{problem}

Adversarial Model Extraction \cite{defazio2019adversarial} presents a framework for model stealing attacks in Graph ML. In this framework, attackers gain access to a small subgraph from the training data of the target model. They then query the target model (e.g., through API access) to gather information about its predictions, denoted as $f_{\text{G}}$, and use this data to train a surrogate model $f_{\text{G}}^*$.
In this scenario, the strategy involves initially generating additional nodes to enhance the accessible subgraph. Subsequently, a GCN model \cite{kipf2017semi} is trained using the augmented subgraph.
In real-world applications, attackers often face incomplete information, such as missing node attributes or incomplete graph topology. A recent taxonomy \cite{wu2022model} categorizes attacker knowledge based on the accessibility of partial node attributes, partial graph topology, and a shadow graph---an auxiliary subgraph from the same domain as the training graph of the target model. Depending on the information available, attackers may need to infer missing attributes or topology before querying the target model $f_{\text{G}}$.
%However, in practice, the information accessible for attackers may be incomplete, e.g., lacking information regarding the node attributes or the graph topology. For example, in the taxonomy presented by \cite{wu2022model}, the authors divide the possible combinations of attacker's knowledge based on the availability of (1) partial node attributes, (2) partial graph topology, and (3) a shadow graph, leading to six different categories. Here both partial node attributes and graph topology refer to the information regarding those nodes to interact with the target model $f_{\text{G}}$, while the shadow graph denotes an auxiliary subgraph from the same domain of the training graph (for the target model). In their proposed strategy, the surrogate model $f_{\text{G}}^*$ can be directly trained by querying a subset of the node labels when the shadow graph is available. When the shadow graph is not available, node attributes or graph topology is first generated before querying $f_{\text{G}}$.
%
% GNNStealing~\cite{shen2022model} further points out that both works above overwhelmingly focus on transductive learning tasks, i.e., the knowledge accessible to the attackers is from the graph data used to train the target model. As this assumption is difficult to hold in practice, GNNStealing proposes a new strategy for model stealing in inductive learning settings. 
GNNStealing \cite{shen2022model} notes that the aforementioned studies primarily focus on transductive learning tasks, where the knowledge accessible to attackers comes from the graph data used to train the target model. Given that this assumption may not always be realistic in practice, GNNStealing introduces a new strategy for model stealing within inductive learning settings.

\subsubsection{Model Stealing Defense}

% Model stealing attacks directly jeopardize the intellectual property and violate the confidentiality of the target model.
%
% We now introduce the methods for the defense of model stealing attacks in the realm of graph machine learning. 
%
% The rationale of model stealing defense is to encode ownership information into an original model, i.e., the target model of model stealing attacks. In this way, the model owner is able to examine whether any suspicious models are stolen ones (i.e., being copied from the original model or being trained directly relying on the knowledge of the original model).
% We formally present the problem definition below.
The rationale behind defending against model stealing is to embed ownership information into the original model—that is, the target model susceptible to model stealing attacks. This approach enables the model owner to determine whether any suspicious models are unauthorized copies or have been trained using knowledge derived from the original model. Below, we formally present the problem definition.
\begin{problem}\label{stealing_defense}
% Given an optimized graph machine learning model $f_{\text{G}}$, the goal of model stealing defense is to achieve a function $h$, such that for any model $f_{\text{G}}^*$ trained relying on $f_{\text{G}}$ and $\tilde{f}_{\text{G}}^*$ trained without relying on $f_{\text{G}}$, we have $h(f_{\text{G}}^*) = 1$ and $h(\tilde{f}_{\text{G}}^*) = 0$.
Given the training graph data $\mathcal{G}$, model stealing defense for Graph ML aims to achieve an optimized original model $f_{\text{G}}$ trained on $\mathcal{G}$, such that for any model $f_{\text{G}}^*$ trained relying on $f_{\text{G}}$ and $\tilde{f}_{\text{G}}^*$ trained without relying on $f_{\text{G}}$, we are able to find a function $h$ where $h(f_{\text{G}}^*) = 1$ and $h(\tilde{f}_{\text{G}}^*) = 0$.
\end{problem}

% To handle this problem, existing works can be mainly divided into two categories, including watermarking and fingerprinting.
To address this issue, existing solutions can be primarily categorized into two types: watermarking and fingerprinting.
%
% Specifically, watermarking methods encode the information of ownership into the original model during the training process, while fingerprinting methods aim to optimize the original model in traditional ways and identify the suspicious model at the inference stage.
%
We elaborate on more details below.

\vspace{0.02in}
\noindent\textbf{Watermarking.}
Watermarking involves embedding secret generated keys into a model during training, allowing the model owner to check for theft by verifying the presence of these keys in suspicious models. 
%The rationale behind watermarking is to generate secret keys and incorporate the information from these keys into the original model during its training phase. This process enables the model owner to determine whether a suspicious model is stolen by checking if it incorporates the secret keys' information. 
%Typically, the information embedded via secret keys does not pertain to any downstream tasks. 
For instance, ER Graph Watermarking \cite{zhao2021watermarking} embeds watermarks into GNNs by creating an Erdős–Rényi random graph with random node features and labels. 
Ownership is confirmed if a suspicious model can correctly classify nodes in this graph.
%Ownership is established by training the GNN model to correctly classify the nodes in this randomly generated graph. To verify a suspicious GNN model, the authors suggest conducting inference using the secret key (i.e., the generated random graph). If the suspicious model can accurately classify most nodes in this graph, it is likely to be a stolen model.
Besides node classification tasks, GNN Watermarking \cite{xu2023watermarking} expands this approach to graph classification tasks by developing a strategy to embed ownership information directly into the training data.

\vspace{0.02in}
\noindent\textbf{Fingerprinting.}
% - model extraction attacks
% Waheed et al.~\cite{waheed2023grove} performed studies to show that 
% In a recent study, GrOVe~\cite{waheed2023grove}, the authors have shown that the output node embedding distributions of two independently trained Graph ML models are different. However, training a model relying on another original model (i.e., producing a stolen model based on an original model) will lead to a large overlap in their output distributions.
In a recent study, GrOVe \cite{waheed2023grove}, researchers demonstrate that the output node embedding distributions of two independently trained Graph ML models differ significantly. However, when one model is trained based on another original model (effectively producing a stolen model), there will be a substantial overlap in their output distributions.
% the output distribution of a model trained relying on an original model will have large overlap with the output distribution of the original model itself.
%
% This reveals the intuition of fingerprinting for Graph ML models, i.e., to differentiate an independently trained model and stolen models by characterizing their output distribution differences. However, existing distribution distance metrics cannot be directly adopted to characterize such differences, since the output distributions of an independently trained model and a stolen model may overlap as well and thus lead to high error rates~\cite{waheed2023grove}.
% %
% Particularly, GrOVe takes an initial step to handle such a problem by first training a set of independently trained models and stolen models. Then a classifier is trained based on such a set of models to identify stolen models based on model output distributions.
%
This highlights the underlying principle of fingerprinting for Graph ML models: distinguishing between independently trained models and stolen models by analyzing differences in their output distributions. However, conventional distribution distance metrics may not effectively characterize these differences because the output distributions of an independently trained model and a stolen model could also overlap, potentially leading to high error rates \cite{waheed2023grove}.
To address this issue, GrOVe takes a preliminary step by first training a set of independently trained models and stolen models. Subsequently, a classifier is developed based on these models to discern stolen models by examining their output distributions.

\vspace{-0.1in}
\section{Applications}

\vspace{0.02in}
\noindent\textbf{Finance.}
Graph ML models are widely used in financial analysis like financial crime detection \cite{suzumura2019towards}. These networks often process sensitive information, including transaction histories and account details \cite{yang2019ffd}. To protect this data, financial institutions implement privacy-preserving measures like differential privacy, which help prevent unauthorized access \cite{fedsage}. Additionally, the financial sector is frequently targeted for various fraudulent activities, including money laundering and identity theft \cite{orekondy2019knockoff}. Therefore, it is essential to develop robust Graph ML models capable of detecting and flagging unusual patterns that may indicate fraudulent activity, thus aiding in crime prevention \cite{hendrycks2021unsolved, dai2022comprehensive}. Moreover, the real-time detection of anomalies within transaction networks is crucial for the timely prevention of financial crimes \cite{jin2021adversarial}.

% %\subsection{Generalizabiltiy}
% \noindent{\textbf{Medical Analysis.}} The application of Graph ML models in medical analysis, such as cancer classification, demonstrates the importance of generalizability in predictions~\cite{wu2022survey}. For instance, Graph ML models can be applied to the task of classifying chest X-ray images into different chest conditions~\cite{mahapatra2022unsupervised}. However, the distribution of images, which are later constructed as graphs, could vary significantly as a result of differences in image-capturing protocols,  devices, or scanner manufacturers. In this case, the generalizability of Graph ML models becomes crucial, as the application scenario would not be consistent during inference.
\vspace{0.02in}
\noindent\textbf{Malware Detection.}
% The widespread use of Internet of Things (IoT) devices has generated extensive graph data, where nodes often represent IoT devices, and edges illustrate their interactions~\cite{yumlembam2022iot}. Malware attacks via smartphone applications pose significant safety risks in these networks, such as unauthorized access to sensitive health information~\cite{busch2021nf}. Employing Graph ML for malware detection can significantly enhance the security and reliability of IoT systems.
The widespread use of Internet of Things (IoT) devices has generated vast amounts of graph data, where nodes typically represent IoT devices and edges depict their interactions \cite{yumlembam2022iot}. Malware attacks through smartphone applications, for instance, pose significant safety risks in these networks, including unauthorized access to sensitive health information \cite{busch2021nf}. Utilizing Graph ML for malware detection can significantly bolster the security and reliability of IoT systems.

%The widespread use of Internet of Things (IoT) devices has produced various forms of graph data~\cite{zhang2022trustworthy}. For example, in IoT networks, nodes can represent individual IoT devices or applications, and edges depict the interactions or communications between them~\cite{yumlembam2022iot}. In these networks, a primary threat to safety is malware attacks through smartphone applications. For instance, an attacker gaining access to a smartphone application can access the sensitive health information of users, posing a vital safety challenge~\cite{busch2021nf}. Here Graph ML could be used to enhance the reliability of the IoT system by malware detection. 
% With techniques of data poisoning defense on graph data~\cite{jin2021adversarial, liu2021graph, yuan2024chasing}, the models could identify abnormal nodes and analyze their intent to detect malware. 

%\subsection{Confidentiality}
\vspace{0.02in}
\noindent\textbf{Healthcare.}
% Graph ML methods are widely applied to healthcare data, where connections between patients based on their shared attributes or medical histories are modeled as graph structures~\cite{pai2018patient,liu2020heterogeneous}.
% %Healthcare data, characterized by complex relationships and interactions, naturally aligns with graph structures~\cite{liu2020heterogeneous}. For instance, patient connections based on shared attributes or medical histories can be graphically modeled to discern treatment response patterns~\cite{pai2018patient}. 
% However, confidentiality remains a critical concern in healthcare, necessitating stringent privacy protections for patient data often dispersed across various institutions~\cite{rajkomar2018scalable, lu2020decentralized}. Federated graph learning addresses this by enabling model training on distributed datasets without compromising individual privacy~\cite{lou2021stfl,bayram2021federated}. Additionally, graph unlearning techniques offer a way to remove data influences from models, thus complying with ethical standards and regulations~\cite{chundawat2023zero, said2023survey}.
Graph ML methods are extensively used in healthcare, where connections among patients, based on shared attributes or medical histories, are represented as graph structures \cite{pai2018patient,liu2020heterogeneous}. However, maintaining confidentiality is a critical issue in healthcare, requiring robust privacy protections for patient data that is often spread across multiple institutions \cite{rajkomar2018scalable, lu2020decentralized}. Federated graph learning provides a solution by allowing model training on distributed datasets without compromising individual privacy \cite{lou2021stfl,bayram2021federated}. Furthermore, graph unlearning techniques enable the removal of data influences from models, ensuring compliance with ethical standards and regulations \cite{chundawat2023zero, said2023survey}.
On the other hand, the generalizability of graph ML models is crucial for applications like cancer classification based on graphs of chest X-ray data~\cite{mahapatra2022unsupervised}. In this case, variations in data distribution due to different devices necessitate consistent performance across distributions.

\vspace{0.02in}
\noindent\textbf{Social Networks.}
On social media platforms such as Reddit and Instagram, data can be modeled as graphs where nodes represent users and edges denote their interactions \cite{li2018ssdmv}. Graph ML models, often used for tasks like content recommendation, are susceptible to significant safety threats. One major concern is the model threat, which involves the potential leakage of users' sensitive information due to inherent message-passing mechanisms that expose data from neighboring nodes \cite{chen2020vertically, olatunji2023releasing}. Implementing differential privacy techniques \cite{sajadmanesh2023gap,wu2022linkteller} can help safeguard privacy by introducing noise to data or model outputs. Additionally, data threats arise from the dynamic and varied nature of social networks, posing challenges to the models' generalizability \cite{wu2022handling}. Employing out-of-distribution (OOD) generalization methods can enhance robustness across diverse social network environments \cite{li2022out}.

\noindent\textbf{Transportation.}
Graph structures are commonly utilized to model dynamic traffic data across transportation networks, leading to the widespread adoption of Graph ML models for tasks like traffic flow prediction \cite{derrow2021eta}. Despite their effectiveness, the generalizability of these models is a critical concern due to the significant variability in traffic conditions and patterns under different circumstances. Thus, developing Graph ML models that perform well across diverse scenarios is crucial for their safe application in predicting traffic flow and congestion \cite{huang2021transfer}. To improve generalizability, domain adaptation techniques are employed to train models that can adapt across various transportation networks \cite{mallick2021transfer}. Additionally, out-of-distribution (OOD) generalization can enhance predictions under unusual circumstances not well represented in the training data, such as during major events or changes in infrastructure \cite{jiang2022graph}.

\section{Future Work}
In this section, we discuss the potential directions for future work of safe Graph ML.

\vspace{0.02in}
\noindent\textbf{Reliability.} Regarding the enhancement of reliability in safety Graph ML, the efficiency of uncertainty quantification is significant yet underexplored. %For aleatory uncertainty, the efficient uncertainty quantification of the 
When enhancing reliability in safety-oriented Graph ML, the effectiveness of uncertainty quantification remains significant but underexplored. For instance, the non-parametric Bayesian GCN \cite{pal2019bayesian} is notably time-consuming due to the Monte Carlo sampling required, making it impractical for high-stakes applications involving large-scale graphs. Additionally, anomaly detection on graphs presents unique challenges; detecting anomalous subgraphs is particularly difficult because the nodes and edges within these subgraphs may appear normal individually, and their structures can vary widely. This complexity has led to a scarcity of research in this specific area \cite{ma2021comprehensive}.

\vspace{0.02in}
\noindent\textbf{Generalizability.} 
Numerous challenges persist in enhancing the generalizability of safe Graph ML. For instance, traditional OOD generalization algorithms typically require specific environmental (i.e., domain) splits during training. However, due to the complex structures inherent in graph data, obtaining such environmental splits can be difficult. It is essential to explore methods that can either manage OOD generalization within a single environment or explicitly divide training environments. Additionally, the approach of test-time training during inference is beneficial for generalizability, as it allows for more flexible use of (unlabeled) data. Developing test-time training tasks that improve model generalizability across various inference scenarios is crucial.

\vspace{0.02in}
\noindent\textbf{Confidentiality.} 
The challenges of safeguarding graph data span multiple dimensions. For instance, there is a need for innovative solutions to facilitate secure collaborative model training across multiple entities while ensuring the confidentiality of node and edge information. Furthermore, existing graph unlearning approaches often lack universality. It is crucial to develop methods capable of addressing all types of unlearning tasks, given the diverse range of applications managed by graph-based models. Additionally, there is a significant gap in evaluating the practicality and scalability of existing methodologies, underscoring the necessity for comprehensive assessments to validate their effectiveness in real-world scenarios.

\section{Conclusion}

In this survey, we conducted a comprehensive review of safety within the rapidly evolving field of graph machine learning (Graph ML), a topic gaining increasing attention due to escalating safety concerns. We provided a structured analysis of three critical safety aspects in Graph ML applications: reliability, generalizability, and confidentiality. To enhance safety, we categorized the threats within each aspect into three main types: data threats, model threats, and attack threats, each presenting unique challenges to Graph ML model safety. For each identified threat, we detailed specific solutions, offering an exhaustive compilation of research efforts addressing these safety challenges. Our discussion synthesized these insights into a unified framework, aimed at deepening the understanding of safety considerations in Graph ML and steering future research in this essential field. Additionally, we highlighted practical applications and suggested directions for future studies. Through this survey, our goal was not only to summarize existing safety research in Graph ML but also to encourage further investigation to ensure that Graph ML techniques are developed and implemented safely.

\label{conclusion_section}

   %\section{Acknowledgments}
%This work is supported by the National Science Foundation under grants (IIS-2006844, IIS-2144209, IIS-2223769, CNS-2154962, BCS-2228534, IIS-2334193, IIS-2217239, CMMI-2146076, CPS-2313110, ECCS-2143559, and ECCS-2033671), the Commonwealth Cyber Initiative awards (VV-1Q23-007, HV2Q23-003, and VV-1Q24-011), the JP Morgan Chase Faculty Research Award, the Cisco Faculty Research Award, and the Department of Energy/Jefferson Lab subcontract 24-D0309.

\linespread{0.95}
\selectfont
\bibliographystyle{plain}
% \bibliography{ijcai21}
\bibliography{ref}

\linespread{1.0}
\selectfont
% \textcolor{red}{author photos are temporarily omitted to facilitate the efficiency of compling.}
% width=1in,height=1.25in
\vspace{-15mm}
\begin{IEEEbiography}[{\includegraphics[width=0.95in,height=1.1875in,clip,keepaspectratio]{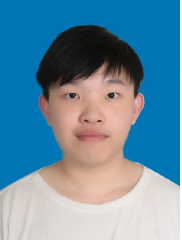}}]{Song Wang}
is a Ph.D. student at the Department of Electrical and Computer Engineering at the University of Virginia, advised by Professor Jundong Li. Previously, he received a B.E. degree in Electronic Engineering from Tsinghua University in 2020. His research interests include knowledge graphs and few-shot learning on graphs. His works have been published in top conferences such as IJCAI, NeurIPS, WSDM, SIGIR, and SIGKDD.
\end{IEEEbiography}
\vspace{-13mm}
\begin{IEEEbiography}[{\includegraphics[width=0.95in,height=1.1875in,clip,keepaspectratio]{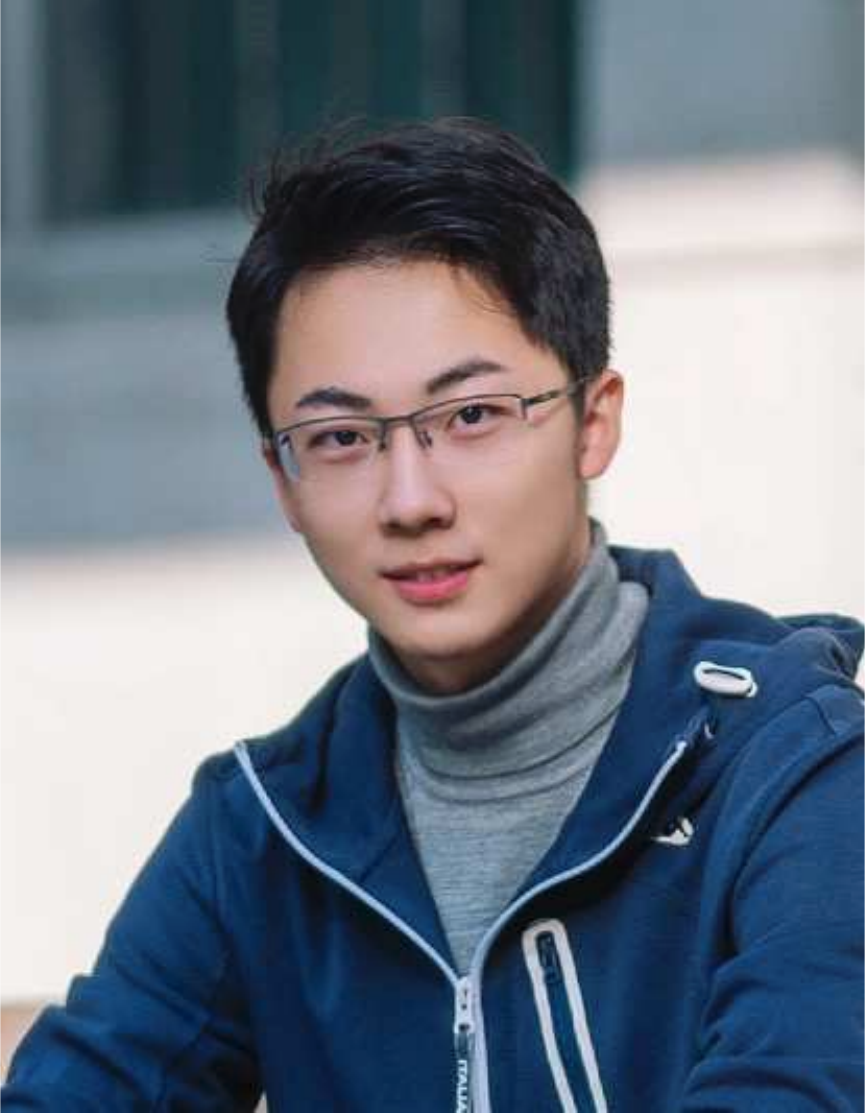}}]{Yushun Dong} is a Ph.D. candidate in the Department of Electrical and Computer Engineering at the University of Virginia. He received a B.S. degree in Telecommunications from Beijing University of Posts and Telecommunications in 2019. His research interests are broadly in data mining and machine learning, with a particular focus on graph learning. In the past few years, his works have been published in top-tier venues including SIGKDD, WWW, and AAAI.
\end{IEEEbiography}
\vspace{-13mm}
\begin{IEEEbiography}[{\includegraphics[width=0.95in,height=1.1875in,clip,keepaspectratio]{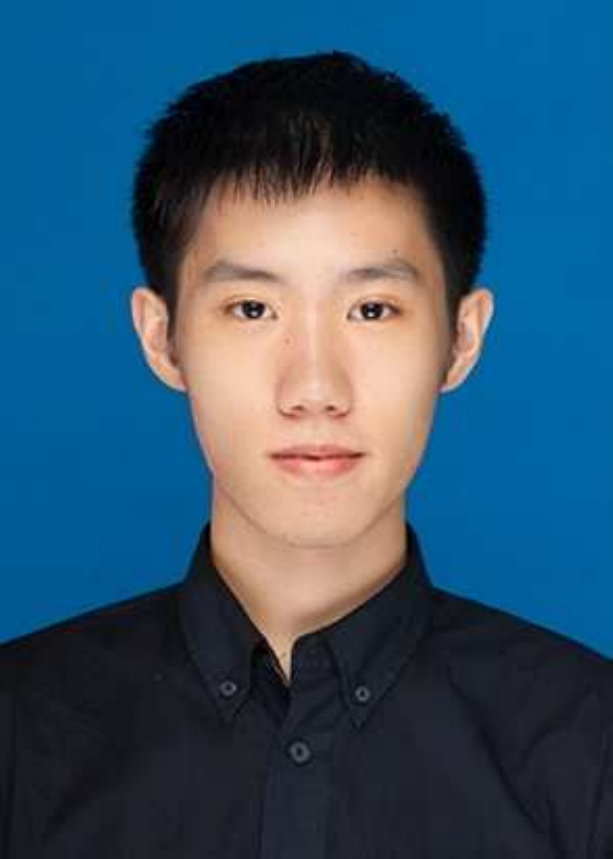}}]{Binchi Zhang} is a Ph.D. student in the Department of Electrical and Computer Engineering at the University of Virginia, advised by Professor Jundong Li. He received his B.E. degree in Electrical Engineering at Xi'an Jiaotong University in 2022. His research interests include graph mining, trustworthy machine learning, and federated learning. His works have been published in top venues including ICLR, SDM, and NeurIPS Datasets and Benchmarks Track.
\end{IEEEbiography}
\vspace{-13mm}
\begin{IEEEbiography}[{\includegraphics[width=0.95in,height=1.1875in,clip,keepaspectratio]{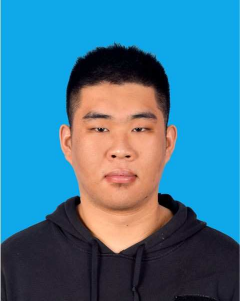}}]{Zihan Chen} is a Ph.D. student in the Department of Electrical and Computer Engineering at the University of Virginia, advised by Professor Jundong Li and Professor Cong Shen. Previously he received his bachelor's degree in mathematics and master's degree in data science from the University of Science and Technology of China. His research interests include federated learning, graph learning, and trustworthy machine learning. 
\end{IEEEbiography}
\vspace{-13mm}
\begin{IEEEbiography}[{\includegraphics[width=0.95in,height=1.1875in,clip,keepaspectratio]{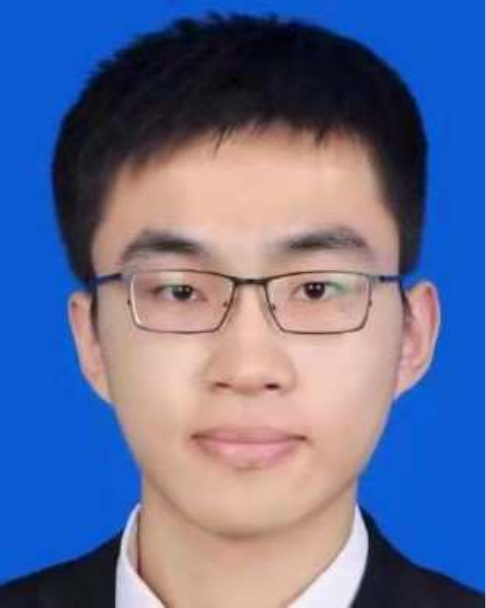}}]{Xingbo Fu} is a Ph.D. student in the Department of Electrical and Computer Engineering at the University of Virginia. He received his B.S. degree and M.S. degree in Automation from Xi'an Jiaotong University in 2017 and 2020, respectively. His research interests include graph learning, federated learning, and AI for healthcare. His recent studies have been published in SIGKDD Explorations, ICHI, and SIGKDD.
\end{IEEEbiography}
\vspace{-13mm}
\begin{IEEEbiography}[{\includegraphics[width=0.95in,height=1.1875in,clip,keepaspectratio]{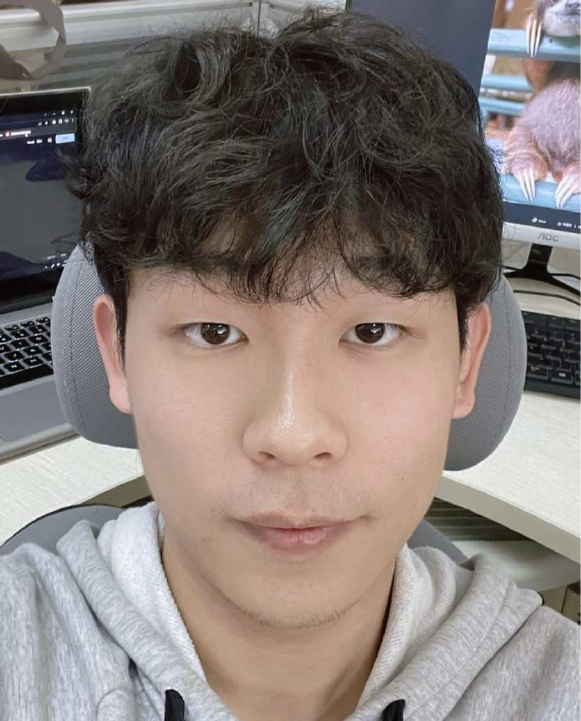}}]{Yinhan He}
 is a Ph.D. student in the Department of Electrical and Computer Engineering at the University of Virginia. He received the B.S. degree in Mathematics and Applied Mathematics from University of Chinese Academy of Sciences in 2022. His research interest is graph machine learning, explainable AI, and machine learning for healthcare.
\end{IEEEbiography}
\vspace{-5mm}
\begin{IEEEbiography}[{\includegraphics[width=0.95in,height=1.1875in,clip,keepaspectratio]{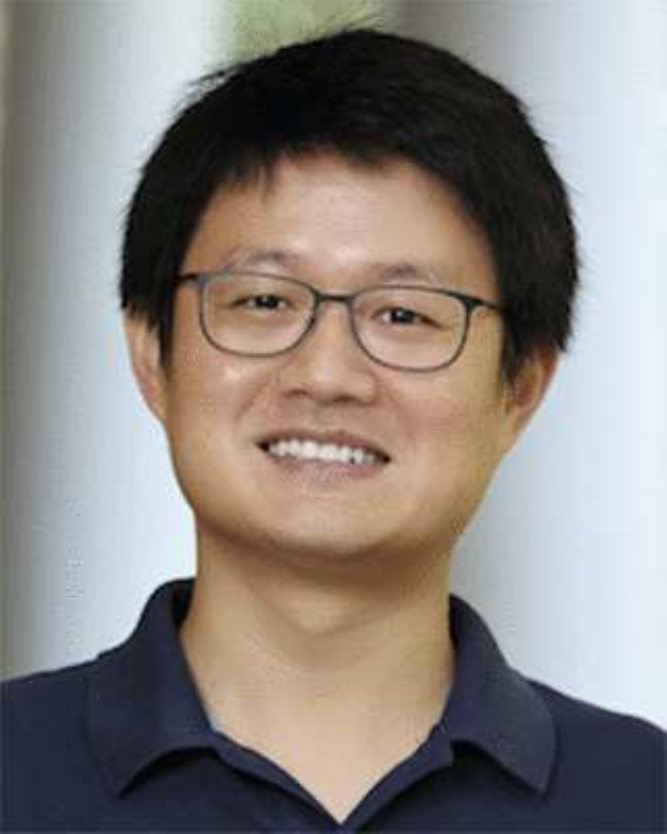}}]{Cong Shen} received the B.S.
and M.S. degrees from the Department of Electronic Engineering, Tsinghua University, China,
in 2002 and 2004, respectively, and the Ph.D.
degree from the Electrical Engineering Department,
UCLA, in 2009. From 2009 to 2014, he worked
for Qualcomm Research, San Diego, CA, USA.
From 2015 to 2019, he was a specially-appointed
Professor with the School of Information Science
and Technology, University of Science and Technology of China (USTC). He is currently an Assistant
Professor with the Charles L. Brown Department of Electrical and Computer
Engineering, University of Virginia. His research interests span a number of
interdisciplinary areas in wireless communications, networking, and machine
learning. He currently serves as an Editor for IEEE Transactions On Wireless Communications and IEEE Wireless Communications Letters.
\end{IEEEbiography}
\vspace{-5mm}
\begin{IEEEbiography}[{\includegraphics[width=0.95in,height=1.1875in,clip,keepaspectratio]{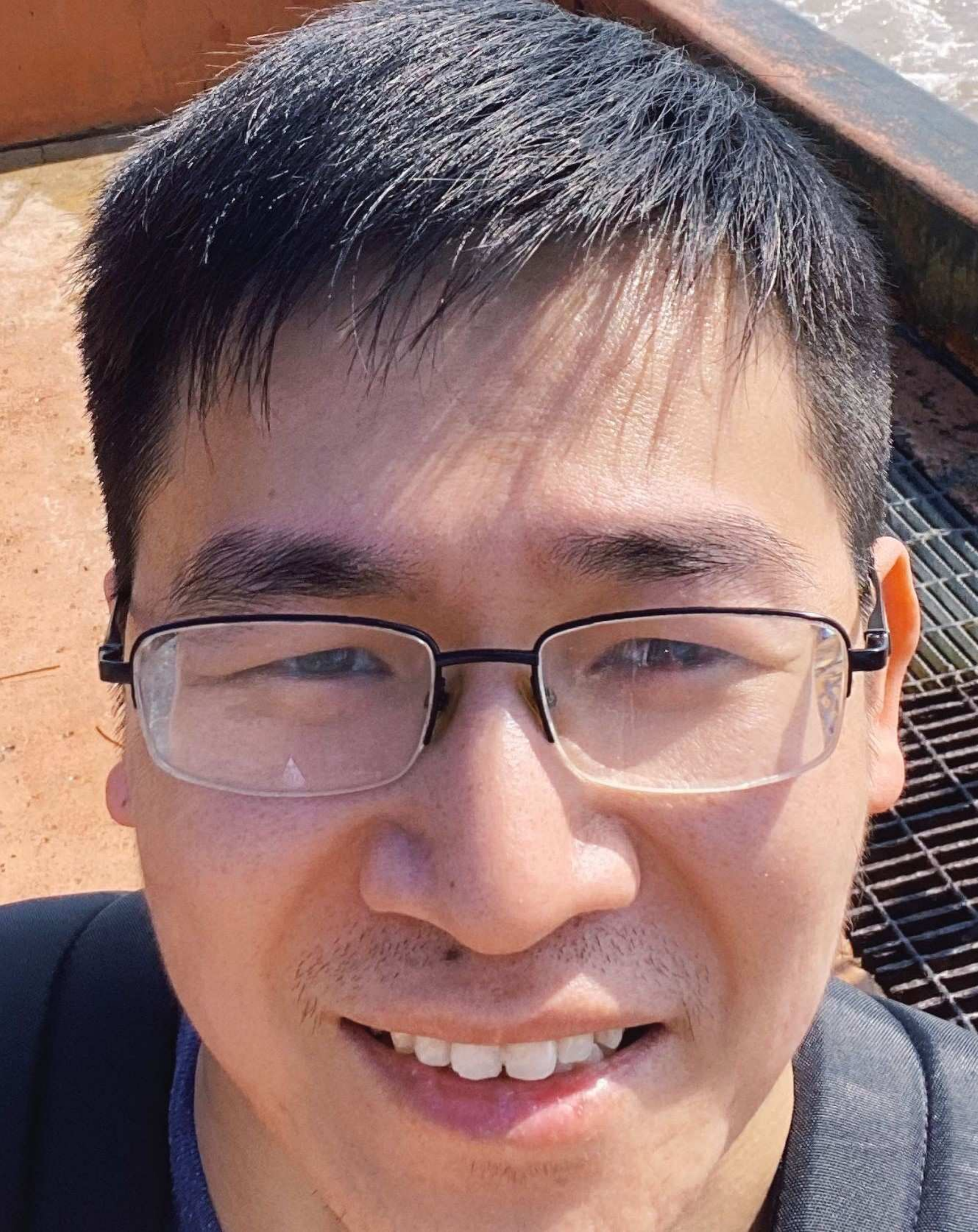}}]{Chuxu Zhang} is an Assistant Professor of Computer Science at the Brandeis University. His research lies in the intersection of artificial intelligence, graph machine learning, and societal applications. Recently, he has focused on developing effective, efficient, safe, and generative machine learning models and algorithms on graph and multi-modality data. Besides, he applies machine learning to solve societal challenges in healthcare, social media, science, and others. His research has led to over 100 papers in major AI venues such as ICML, NeurIPS, ICLR, KDD, WWW, and EMNLP. He is the recipient of the NSF CAREER Award (2024) and several best paper (candidate) awards including CIKM 2021 and WWW 2022. 
\end{IEEEbiography}
\vspace{-5mm}
\begin{IEEEbiography}[{\includegraphics[width=0.95in,height=1.1875in,clip,keepaspectratio]{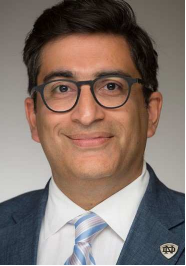}}]{Nitesh V. Chawla} (Fellow, IEEE) received the PhD
degree in computer science and engineering from
the University of South Florida, Tampa, Florida, in
2002. He is the Frank M. Freimann professor of
computer science and engineering with the University of Notre Dame. He is the founding director of
the Lucy Family Institute for Data and Society. His
research interests focus on machine
learning, data science, and network science, and is
motivated by the question of how technology can
advance the common good through interdisciplinary research.
\end{IEEEbiography}
\vspace{-5mm}
\begin{IEEEbiography}[{\includegraphics[width=0.95in,height=1.1875in,clip,keepaspectratio]{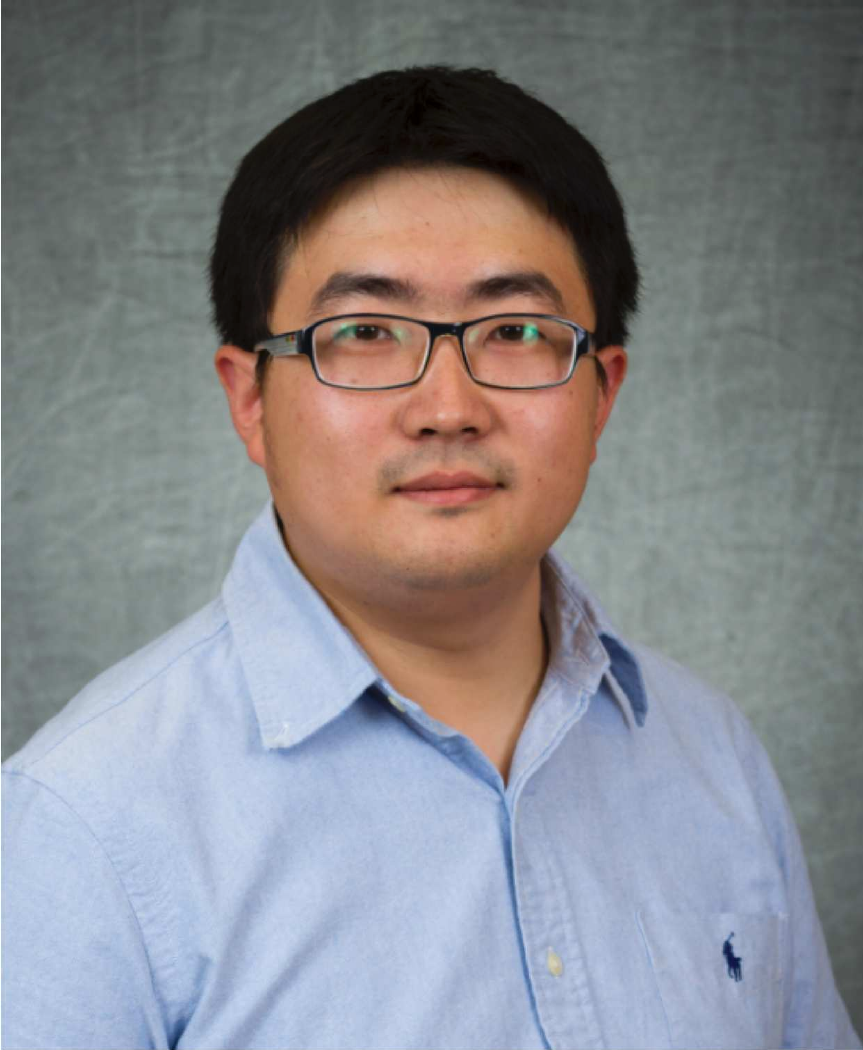}}]{Jundong Li} is an Assistant Professor in the Charles L. Brown Department of Electrical and Computer Engineering, with a joint appointment in the Department of Computer Science, and the School of Data Science. He received his Ph.D. degree in Computer Science at Arizona State University in 2019. His research interests are in data mining, machine learning, and causal inference. He has published over 150 articles in high-impact venues (e.g., KDD, WWW, NeurIPS, ICLR, AAAI, IJCAI, WSDM, EMNLP, CSUR, TPAMI, TKDE, TKDD, and TIST). He has won several prestigious awards, including NSF CAREER Award, KDD Best Research Paper Award, JP Morgan Chase Faculty Research Award, Cisco Faculty Research Award, among others.
\end{IEEEbiography}

\clearpage

%\appendices

\end{document}

%% file: category.tex
\usetikzlibrary{trees}
\usetikzlibrary{matrix, patterns, positioning}

\definecolor{connect-line}{RGB}{0,0,0}
\definecolor{middle-color}{RGB}{255,255,255}
% \definecolor{leaf-color}{RGB}{166,208,153}
% \definecolor{leaf-color}{RGB}{173,216,230}
\definecolor{leaf-color}{RGB}{255,255,255}
% \definecolor{line-color}{RGB}{166,208,153}
\definecolor{line-color}{RGB}{25,25,112}

\definecolor{black}{RGB}{0,0,0}

% \definecolor{reliability}{RGB}{230,75,53}
% \definecolor{generalizability}{RGB}{7,105,82}
% \definecolor{confidentiality}{RGB}{207,134,0}

\definecolor{reliability}{RGB}{240, 128, 128}
\definecolor{generalizability}{RGB}{124, 205, 124}
\definecolor{confidentiality}{RGB}{135, 206, 235}

\tikzstyle{reliability-leaf}=[draw=reliability,
    rounded corners,minimum height=1em,
    fill=leaf-color!40,text opacity=1, align=left,
    fill opacity=.5,  text=black,align=left,font=\scriptsize,
    inner xsep=3pt,
    inner ysep=1pt,
]
\tikzstyle{reliability-middle}=[draw=reliability,
    rounded corners,minimum height=1em,
    fill=middle-color!40,text opacity=1, align=center,
    fill opacity=.5,  text=black,align=center,font=\scriptsize,
    inner xsep=3pt,
    inner ysep=1pt,
]
    
\tikzstyle{generalizability-leaf}=[draw=generalizability,
    rounded corners,minimum height=1em,
    fill=leaf-color!40,text opacity=1, align=left,
    fill opacity=.5,  text=black,align=left,font=\scriptsize,
    inner xsep=3pt,
    inner ysep=1pt,
]
\tikzstyle{generalizability-middle}=[draw=generalizability,
    rounded corners,minimum height=1em,
    fill=middle-color!40,text opacity=1, align=center,
    fill opacity=.5,  text=black,align=center,font=\scriptsize,
    inner xsep=3pt,
    inner ysep=1pt,
]

\tikzstyle{confidentiality-leaf}=[draw=confidentiality,
    rounded corners,minimum height=1em,
    fill=leaf-color!40,text opacity=1, align=left,
    fill opacity=.5,  text=black,align=left,font=\scriptsize,
    inner xsep=3pt,
    inner ysep=1pt,
]
\tikzstyle{confidentiality-middle}=[draw=confidentiality,
    rounded corners,minimum height=1em,
    fill=middle-color!40,text opacity=1, align=center,
    fill opacity=.5,  text=black,align=center,font=\scriptsize,
    inner xsep=3pt,
    inner ysep=1pt,
]

\tikzstyle{leaf}=[draw=line-color,
    rounded corners,minimum height=1em,
    fill=leaf-color!40,text opacity=1, align=center,
    fill opacity=.5,  text=black,align=center,font=\scriptsize,
    inner xsep=3pt,
    inner ysep=1pt,
    ]
\tikzstyle{middle}=[draw=line-color,
    rounded corners,minimum height=1em,
    fill=middle-color!40,text opacity=1, align=center,
    fill opacity=.5,  text=black,align=center,font=\scriptsize,
    inner xsep=3pt,
    inner ysep=1pt,
    ]

\forestset{
  custom edge/.style={
edge path={\noexpand\path[\forestoption{edge}, rounded corners](!u.parent anchor) -- (.child anchor)\forestoption{edge label}}
  }
}
\forestset{
  straight edge/.style={
    edge path={
      \noexpand\path [\forestoption{edge}] (!u.parent anchor) -- (.child anchor)\forestoption{edge label};
    }
  }
}

\begin{figure*}[ht]
\centering
\begin{forest}
  for tree={
edge={-, draw=connect-line, line width=1pt},
  edge path={
    \noexpand\path[\forestoption{edge}, rounded corners]
    (!u.parent anchor) -- +(-1pt,0pt)  -| ([xshift=0pt].child anchor)\forestoption{edge label};
  },
    grow=east,
    reversed=true,
    anchor=base west,
    parent anchor=east,
    child anchor=west,
    base=middle,
    font=\scriptsize,
    rectangle,
    line width=1.2pt,
    draw=connect-line,
    rounded corners,align=left,
    minimum width=2em,
    s sep=4pt,
    inner xsep=3pt,
    inner ysep=0pt,
  },
  where level=1{text width=4.5em}{},
  where level=2{text width=5em,font=\scriptsize}{},
  where level=3{font=\scriptsize}{},
  where level=4{straight edge, font=\scriptsize}{},
  where level=5{font=\scriptsize}{},
  [\textbf{Safety in }  \\ \textbf{Graph ML} ,  anchor=north,edge=reliability, 
    [\textbf{Reliability}, reliability-middle, edge=reliability, text width=5.5em
        [Model Threat: Uncertainty, reliability-middle, edge=reliability, text width=10em,
            [Aleatory Uncertainty, reliability-middle, text width=8em, edge=reliability
                [Bayes-NC\cite{dallachiesa2014node}{,}
                URGE~\cite{hu2017embedding}{,}
                Bayesian GCN~\cite{zhang2019bayesian}{,}\\
                BGCN~\cite{pal2019bayesian}{,}
                GGP~\cite{ng2018bayesian}
, reliability-leaf, text width=18.0em,  edge=reliability ]
            ]
            [Episdemic Uncertainty, reliability-middle, text width=8em, edge=reliability
                [GCN-BBGDC~\cite{hasanzadeh2020bayesian}{,}
                STZINB-GNN~\cite{zhuang2022uncertainty}{,}
                CF-GNN~\cite{huang2023uncertainty}, reliability-leaf, text width=18.0em, edge=reliability]
            ]
        ]
        [Data Threat: Anomalies, reliability-middle, edge=reliability, text width=10em, pattern=crosshatch, pattern color=reliability
            [In-Graph Anomalies, reliability-middle, text width=8em, edge=reliability,pattern=crosshatch, pattern color=reliability
                [Radar \cite{li2017radar}{,}
DOMINANT~\cite{ding2019deep}{,} CARE~\cite{dou2020enhancing}{,}  \\
ANOMALOUS~\cite{peng2018anomalous} {,}
 PC-GNN~\cite{liu2021pick}{,}
%$\mathrm{H^{2}}$-FDetector~\cite{shi2022h2}{,}
CONAD~\cite{xu2022contrastive}{,} \\
GCCAD~\cite{chen2022gccad}{,} 
  CoLA~\cite{liu2021anomaly}{,} DONE~\cite{bandyopadhyay2020outlier}{,} BWGNN~\cite{tang2022rethinking}{,}\\
  Sub-CR~\cite{zhang2022reconstruction}{,} SL-GAD~\cite{zheng2021generative} {,} ACT~\cite{wang2023cross}{,}
,reliability-leaf, text width=18.0em, edge=reliability,] %ACT~\cite{wang2023cross}{,} OCGNN~\cite{wang2021one}{,}
            ]
            [Cross-Graph Anomalies, reliability-middle, text width=8em, edge=reliability,pattern=crosshatch, pattern color=reliability
                [GAWD~\cite{lee2021gawd}{,} M$^3$FEND~\cite{zhu2022memory}{,} UPFD~\cite{dou2021user}{,} \\
                OCGIN~\cite{zhao2023using}{,} DeepSphere~\cite{teng2018deep}{,} GLAD-PAW~\cite{wan2021glad}
                , reliability-leaf, text width=18.0em, edge=reliability]
            ]
        ]
        [Attack Threat: Poisoning, reliability-middle, edge=reliability, text width=10em, fill=reliability!40
            [Data Poisoning Attack, reliability-middle, text width=8em, edge=reliability,fill=reliability!40
                [\cite{bojchevski2019adversarial}{,} CD-Attack~\cite{li2020adversarial}{,} \cite{xu2019topology}{,} Opt-Attack~\cite{sun2018data}{,} \\
                \cite{bhardwaj2021adversarial}{,} \cite{bhardwaj2021poisoning}{,} \cite{wang2019attacking}{,} Nettack~\cite{zugner2018adversarial}{,} Metattack~\cite{zugner2020adversarial}{,} \\
                AtkSE~\cite{liu2022gradients}{,} NIPA~\cite{sun2019node}{,} G-FairAttack~\cite{zhang2023adversarial}{,} \\
                FATE~\cite{kang2023deceptive}{,} Graph-Fraudster~\cite{chen2022graphfraudster}
                , reliability-leaf, text width=18.0em, edge=reliability ] % \cite{xu2022more} 
            ]
            %edge path={\noexpand\path[\forestoption{edge}, rounded corners](!u.parent anchor) -- (.child anchor)}
            [Data Poisoning Defense, reliability-middle, text width=8em, edge=reliability,fill=reliability!40
                [\cite{zhang2021detection}{,} NeuralSparse~\cite{zheng2020robust}{,} \cite{entezari2020all}{,} ProGNN~\cite{jin2021adversarial}{,} \\
                LRGNN~\cite{xu2021speedup}{,} \cite{tang2020transferring}{,} SimPGCN~\cite{jin2021node}{,} GAME~\cite{zhang2023chasing}{,}\\ AirGNN~\cite{liu2021graph}{,} Dragon~\cite{yuan2024chasing}{,}
                GCN-LFR~\cite{chang2021not}
                , reliability-leaf, text width=18.0em, edge=reliability]
            ]
        ]
    ]
    [\textbf{Generalizability}, generalizability-middle, edge=generalizability, text width=5.5em
        [Model Threat: Adaptation, generalizability-middle, edge=generalizability,text width=10em, 
            [Domain Adaptation, generalizability-middle, text width=8em, edge=generalizability
                [DANE~\cite{zhang2019dane}{,} DGDA~\cite{cai2021graph}{,}
                {GraphAE}~\cite{guo2022learning}{,}
{GRADE}~\cite{wu2023non}{,}
\\ %GCAN~\cite{ma2019gcan}{,}
{SGDA}~\cite{qiao2023semi}{,}
%FD-GCN~\cite{mahapatra2022unsupervised}{,} 
UDA-GCN~\cite{wu2020unsupervised}{,}
AdapterGNN~\cite{li2023adaptergnn} , generalizability-leaf, text width=18.0em, edge=generalizability]
            ]
            [Test-Time Adaptation, generalizability-middle, text width=8em, edge=generalizability                
                [%AUX-TS~\cite{han2021adaptive}{,} AdaGCN~\cite{dai2022graph}{,} EGI~\cite{zhu2021transfer}{,}\\ 
                %TL-DCRNN~\cite{mallick2021transfer}{,} TEEPEE~\cite{huang2021transfer}
                {SOGA}~\cite{mao2024source}{,}
{GAPGC}~\cite{chen2022graphtta}{,}
{GT3}~\cite{wang2022test}{,}
{G-GLOW}~\cite{zhao2023graphglow}
, generalizability-leaf, text width=18.0em, edge=generalizability]
            ]
        ]
        [Data Threat: Unseen Data, generalizability-middle, edge=generalizability, text width=10em,
        pattern=crosshatch, pattern color=generalizability
            [Adversarial Training, generalizability-middle, text width=8em, edge=generalizability,pattern=crosshatch, pattern color=generalizability
                    [ CAP~\cite{xue2021cap}{,}
GraphAT~\cite{feng2019graph}{,} {DAGNN}~\cite{wu2019domain}{,}\\
GNN-DRO~\cite{sadeghi2021distributionally}{,} WT-AWP~\cite{wu2023adversarial}, generalizability-leaf, text width=18.0em, edge=generalizability]
            ]
            [Invariant Learning, generalizability-middle, text width=8em, edge=generalizability,pattern=crosshatch, pattern color=generalizability
                [SR-GNN~\cite{zhu2021shift}{,} EERM~\cite{wu2022handling}{,} DIR~\cite{wu2022discovering}{,}
GIL~\cite{li2022learning}{,} \\
CAL~\cite{sui2022causal}{,} E-invariant GR~\cite{bevilacqua2021size}{,}
CIGA~\cite{chen2022learning}{,}\\
GSAT~\cite{miao2022interpretable}{,} IS-GIB~\cite{yang2023individual}{,} MARIO~\cite{zhu2023mario}{,} LiSA~\cite{yu2023mind}, generalizability-leaf, text width=18.0em, edge=generalizability]
            ]
        ]
        [Attack Threat: Evasion, generalizability-middle, edge=generalizability, text width=10em,
        fill=generalizability!40
            [Model Evasion Attack, generalizability-middle, text width=8em, edge=generalizability,fill=generalizability!40
                [\cite{chen2017practical}{,} FGA~\cite{chen2018fast}{,} \cite{lin2020adversarial}{,} RL-S2V~\cite{dai2018adversarial}{,} Q-Attack~\cite{chen2019ga}{,} \\
                IG-JSMA~\cite{wu2019adversarial}{,} \cite{zhang2021projective}{,} GF-Attack~\cite{chang2020restricted}{,} RWCS~\cite{ma2020towards}{,} \\
                \cite{mu2021hard}{,} GRABNEL~\cite{wan2021adversarial}{,} FA-GNN~\cite{hussain2022adversarial} 
                , generalizability-leaf, text width=18.0em, edge=generalizability]
            ]
            [Model Evasion Defense, generalizability-middle, text width=8em, edge=generalizability,fill=generalizability!40
                [\cite{xu2019topology}{,} \cite{dai2019adversarial}{,}   LAT-GCN~\cite{jin2019latent}{,} GraphDefense~\cite{wang2019graphdefense}{,} % GraphVAT~\cite{feng2019graph}{,} \cite{zhou2019adversarial}{,}\cite{carlini2017adversarial}{,} 
                \cite{bojchevski2019certifiable}{,} \\\cite{zugner2019certifiable}{,} %\cite{xu2018characterizing}{,} \cite{zhang2019comparing}{,}
                ELEGANT~\cite{dong2023elegant}{,} GraphSAC~\cite{ioannidis2019graphsac}{,} % \cite{jia2020certified}{,} 
               VPN~\cite{jin2019power}
,generalizability-leaf, text width= 18.0em, edge=generalizability]
            ]
        ]
    ]
    [\textbf{Confidentiality}, confidentiality-middle, edge=confidentiality, text width=5.5em
        [Model Threat: Privacy, confidentiality-middle, edge=confidentiality, text width=10em
            [Differential Privacy, confidentiality-middle, text width=8em, edge=confidentiality
                [VFGNN~\cite{chen2020vertically}{,} DP-GNN~\cite{daigavane2021node}{,} LPGNN~\cite{sajadmanesh2021locally}{,} \\
                PrivGnn~\cite{olatunji2023releasing}{,} GDP~\cite{chien2023differentially}{,} PPGD~\cite{zhang2019graph}{,} \\
                DPNE~\cite{xu2018dpne}{,} GAP~\cite{sajadmanesh2023gap}{,} Linkteller~\cite{wu2022linkteller}{,} SGNN~\cite{mei2019sgnn} , confidentiality-leaf, text width=18.0em, edge=confidentiality]
            ]
            [Graph Unlearning, confidentiality-middle, text width=8em, edge=confidentiality
                [GraphEditor~\cite{cong2022grapheditor}{,} GraphEraser~\cite{chen2022graph}{,} SGCunlearn\cite{chien2023efficient}{,}\\
                GNNDelete~\cite{cheng2023gnndelete}{,}  GIF~\cite{wu2023gif}{,}  CEU~\cite{wu2023certified}{,} GST~\cite{pan2023unlearning},confidentiality-leaf, text width= 18.0em, edge=confidentiality]
            ]
        ]
        [Data Threat: Distributed Data, confidentiality-middle, edge=confidentiality, text width=10em,         pattern=crosshatch, pattern color=confidentiality
            [Data Heterogeneity, confidentiality-middle, text width=8em, edge=confidentiality,pattern=crosshatch, pattern color=confidentiality
                [FedPub~\cite{fedpub}{,} FGSSL~\cite{huang2023federated}{,} FedCog~\cite{lei2023federated}{,}
                FedSage~\cite{fedsage}{,}\\
                GCFL~\cite{xie2021gcfl}{,} FedLit~\cite{fedlit}{,} 
                FedStar~\cite{fedstar}{,} 
                FLIT~\cite{zhu2022flitplus}{,} \\
                STFL~\cite{lou2021stfl}{,} 
                GraphFL~\cite{wang2022graphfl}{,} FedGCN~\cite{yao2022fedgcn},confidentiality-leaf, text width= 18.0em, edge=confidentiality]
            ]
            [Overlapping Instances, confidentiality-middle, text width=8em, edge=confidentiality,pattern=crosshatch, pattern color=confidentiality
                [FedE~\cite{chen2021fede}{,} FedR~\cite{zhang2022fedr}{,} FKGE~\cite{peng2021fkge}{,} \\FeSoG~\cite{liu2021fesog}{,} FedPerGNN~\cite{wu2022fedpergnn}
                ,confidentiality-leaf, text width= 18.0em, edge=confidentiality]
            ]
        ]
        [Attack Threat: Stealing, confidentiality-middle, edge=confidentiality, text width=10em,
fill=confidentiality!40        
            [Model Stealing Attack, confidentiality-middle, text width=8em, edge=confidentiality,fill=confidentiality!40    
                [Adversarial Model Extraction \cite{defazio2019adversarial}{,} Six-type \\ Model Extraction Attacks \cite{wu2022model}{,} GNNStealing~\cite{shen2022model}, confidentiality-leaf, text width=18.0em, edge=confidentiality]
            ]
            [Model Stealing Defense, confidentiality-middle, text width=8em, edge=confidentiality,fill=confidentiality!40    
                [ER Graph Watermarking \cite{zhao2021watermarking}{,}  GNN \\Watermarking \cite{xu2023watermarking}{,} GrOVe \cite{waheed2023grove} ,confidentiality-leaf, text width= 18.0em, edge=confidentiality]
            ]
        ]
    ]
  ]  
\end{forest}
\vspace{-0mm}
\caption{A taxonomy of safety problems and solutions on Graph ML and the related works.}
\vspace{-3mm}
\label{fig:taxonomy-techniques}
\end{figure*}

%% file: generalizability_model.tex
\subsection{Model Threat: Adaptation}
% The crucial part of adaptation is to transfer the learned knowledge from labeled data (i.e., the source distributions) during training to the unlabeled data (i.e., the target distributions)~\cite{li2023adaptergnn,dai2022graph}. 
A crucial aspect of adaptation involves transferring the knowledge acquired from labeled data (i.e., the source distributions) during training to unlabeled data (i.e., the target distributions) \cite{li2023adaptergnn, dai2022graph}.
%Graph ML models, typically tailored for specific objectives during training, can struggle to adapt when faced with new distributions during inference~\cite{li2023adaptergnn,dai2022graph}. 
%This issue presents a threat to the generalizability of graph ML models as they could encounter diverse application scenarios requiring adaptability to a range of datasets and tasks~\cite{huang2021transfer}.
% For example, in tumor classification, X-ray images are often represented as graphs by connecting regions of pixels. Since these graphs may originate from multiple medical centers using various devices, significant differences in data distribution can occur~\cite{mahapatra2022unsupervised}. In such cases, the model must effectively learn from labeled data at one center and apply this knowledge to classify unlabeled graphs from another center, to maintain diagnostic accuracy across data from diverse devices.
%As a result, it becomes crucial to enhance the generalizability of graph ML models by adapting to various data distributions. Note that in this subsection, we focus on adaption tasks where the data for inference is accessible during training, and we will discuss the scenario of handling unseen data during training in the next subsection.
For instance, in tumor classification, X-ray images are often represented as graphs by connecting regions of pixels. Given that these graphs may come from multiple medical centers using different devices, significant variations in data distribution can occur \cite{mahapatra2022unsupervised}. In these scenarios, it is crucial for the model to effectively learn from labeled data at one center and then apply this knowledge to classify unlabeled graphs from another center, ensuring diagnostic accuracy across data from diverse devices.
%

% Below we introduce safeguard techniques to enhance generalizability via adaptation.
% Particularly, we introduce two types of approaches that extend models' learned knowledge and help achieve greater flexibility in handling multiple data distributions: \textit{graph domain adaptation} and \textit{test-time adaptation}. 
Below, we discuss safeguard techniques designed to enhance generalizability through adaptation. Specifically, we explore two approaches that extend the models' learned knowledge and provide flexibility in handling various data distributions: \textit{graph domain adaptation} and \textit{test-time adaptation}.
%Generally, graph domain adaptation aims to leverage the rich information in existing labeled graphs and adapt the model to unlabeled ones during training~\cite{zhang2019dane,mahapatra2022unsupervised}, while test-time adaptation aims to transfer knowledge in a trained model to downstream data distributions without access to the labeled data during training.

\subsubsection{Graph Domain Adaptation}
%The task of graph domain adaptation is to adapt the model to various data distributions during training. 
% Given a source graph distribution with fully labeled data and a target unlabeled graph distribution, the goal of graph domain adaptation is to leverage the abundant labeled information in the source graph distribution and adapt the learned knowledge to the target graph distribution~\cite{wu2020unsupervised}. 
In graph domain adaptation, the objective is to utilize the abundant labeled information from a source graph distribution and transfer the learned knowledge to an unlabeled target graph distribution. This process aims to bridge the gap between the two distinct distributions, allowing for effective application across different data settings \cite{wu2020unsupervised}.
%Thus, the most crucial part of graph domain adaptation is to maximally learn the distribution of unlabeled data, such that the supervision information in labeled data could be adapted to unlabeled data. 

\vspace{0.02in}
\noindent\textbf{Distribution Alignment.}
% A prevalent solution for graph domain adaptation is to align the source distribution with the target distribution. With this concept, a series of studies are conducted on different distribution discrepancy measurements~\cite{li2022out}.
A common approach to graph domain adaptation involves aligning the source distribution with the target distribution. This strategy has led to a series of studies focused on various methods for measuring distribution discrepancies \cite{li2022out}.
%
% SR-GNN~\cite{zhu2021shift} leverages central moment discrepancy as the measurements, acting as a regularization term for model optimization. SR-GNN focuses on node representations learned by the last activated hidden layers of GNN models to maximize the effect of the regularization of distribution alignment. More recently, GraphAE~\cite{guo2022learning} also proposes to minimize distribution discrepancy, except that GRADE leverages message routing alignment and message aggregating alignment to reduce the difference (between the source distribution and the target distribution) in terms of structures and node features, respectively. These two alignments are applied to model optimization based on the proposed distance loss in the form of the multiple kernel variant of maximum mean discrepancy (MK-MMD). Furthermore, based on the message-passing mechanism in GNNs~\cite{kipf2017semi}, GRADE~\cite{wu2023non} further proposes a novel discrepancy measurement named Graph Subtree Discrepancy, which estimates the similarity of subtree representations learned by GNNs. In this manner, the discrepancy measurement could benefit from the expressiveness of GNNs.
SR-GNN \cite{zhu2021shift} utilizes central moment discrepancy as a metric, serving as a regularization term during model optimization. It specifically focuses on node representations learned by the last activated hidden layers of GNN models to maximize the effectiveness of distribution alignment regularization. Building on similar concepts, GraphAE \cite{guo2022learning} also aims to minimize distribution discrepancies. However, GraphAE employs message routing alignment and message aggregating alignment to address differences in structures and node features, respectively, between the source and target distributions. These alignments are incorporated into model optimization through a proposed distance loss using a multiple kernel variant of maximum mean discrepancy (MK-MMD). Furthermore, leveraging the message-passing mechanism typical of GNNs \cite{hamilton2017inductive}, GRADE \cite{wu2023non} introduces a novel discrepancy metric called Graph Subtree Discrepancy, which assesses the similarity of subtree representations generated by GNNs. This approach allows the discrepancy measurement to benefit from the expressive power of GNNs.

%to leverage self-supervised learning strategies such as contrastive learning~\cite{oord2018representation,he2020momentum}, as they are more suitable for learning from unlabeled data. For example, DGDA~\cite{cai2021graph} designs a disentanglement-based algorithm for graph-structured data. This method employs variational graph auto-encoders~\cite{kipf2016variational,simonovsky2018graphvae} to extract latent variables and then disentangle them using three distinct supervised learning modules. These modules are designed to force the model to distinguish between different domains, label distributions, and random noise. The combination of self-supervised and supervised learning effectively extracts the patterns of unlabeled data and enhances model adaptability.
\vspace{0.02in}
\noindent\textbf{Adversral Training.}
% Another line of work focuses on an adversarial training manner to perform domain adaptation. Generally, these works incorporate a distribution discriminator to learn representations from graph ML models that could confuse the discriminator. As a result, the learned representations become more robust to the variance among distributions.
Another approach in domain adaptation employs adversarial training techniques. Typically, this involves integrating a distribution discriminator designed to differentiate between source and target distributions. By training Graph ML models to produce representations that can confuse this discriminator, the resulting representations become more robust and less sensitive to variations across different distributions.
%
% \textit{DAGNN}~\cite{wu2019domain} aims to learn generalizable graph representations that are difficult to classify by the distribution discriminator, while involving information that is decisive for predictions by the label classifier. Nevertheless, the optimization in DAGNN is iteratively performed between the distribution discriminator and the label classifier, which could be less effective. To bridge this gap, \textit{DANE}~\cite{zhang2019dane} instead proposes an adversarial loss as a regularization term that aligns representations learned from different distributions. By introducing the loss to both the distribution discriminator and the label classifier, the optimization of these two modules could be achieved simultaneously. 
% %
% Furthermore, to gain a global view of both source and target distributions, \textit{UDA-GCN}~\cite{wu2020unsupervised} learns representations while considering the global consistency of graphs among distributions. Notably, UDA-GCN employs different classifiers for the source and target distributions, in order to achieve adaptation to both distributions.
% %
% More recently, to deal with the potential label scarcity problem in the source distribution, \textit{SGDA}~\cite{qiao2023semi} additionally performs pseudo-labeling for unlabeled nodes based on their overall distance to nodes in each class.
DAGNN \cite{wu2019domain} aims to develop generalizable graph representations that are indistinguishable by a distribution discriminator, while still retaining crucial information for prediction tasks as determined by the label classifier. However, the optimization in DAGNN, which iteratively alternates between tuning the distribution discriminator and the label classifier, can be suboptimal. To address this issue, DANE \cite{zhang2019dane} introduces an adversarial loss as a regularization term to align representations learned from different distributions. This allows for simultaneous optimization of both the distribution discriminator and the label classifier.
Furthermore, to encompass a global perspective of both source and target distributions, UDA-GCN \cite{wu2020unsupervised} learns representations while ensuring global consistency across distributions. Uniquely, UDA-GCN utilizes separate classifiers for the source and target distributions to facilitate adaptation to both.
More recently, in response to potential label scarcity in the source distribution, SGDA \cite{qiao2023semi} incorporates pseudo-labeling for unlabeled nodes by assessing their distances to nodes in each class, enhancing the model's effectiveness under limited label availability.

%For example, GCAN~\cite{ma2019gcan} proposes to jointly learn from data structures, domain labels, and classes in a unified framework. In alignment with this idea, the authors design three effective alignment mechanisms, including structure alignment, domain alignment, and class alignment, to learn domain-invariant representations. During this process, an adversarial loss is introduced to enforce the model to distinguish between different domains. 
%UDA-GCN~\cite{wu2020unsupervised} explores the problem of cross-domain node classification. The framework first captures the local and global consistency relationship of each graph and combines them with an attention-based mechanism. After that, the framework utilizes three classifiers designed to collaboratively learn domain-invariant and semantic representations. These classifiers are specifically tailored for training the source classifier, the domain classifier, and the target classifier, respectively. 

\subsubsection{Test-Time Adaptation}
% Test-time adaptation focuses on generalizing a pre-trained model (on the source distribution) to the target distribution, without access to the source distribution during adaptation. Test-time adaptation overcomes the limitations of traditional domain adaptation that requires both source and target distributions for adpataion~\cite{liu2024beyond}.%are required but may not be accessible in practice, due to privacy concerns~\cite{liu2024beyond}.
%
Test-time adaptation aims to generalize a pre-trained model (developed on the source distribution) to the target distribution, without requiring access to the source distribution during adaptation. This method overcomes the constraints of traditional domain adaptation, which typically necessitates the presence of both source and target distributions for effective adaptation \cite{liu2024beyond}.

\vspace{0.02in}
\noindent\textbf{Full Fine-tuning.} 
%Given a model pre-trained on the source distribution, an intuitive strategy for test-time adaptation is to fine-tune the entire model on the target distribution. As such, it becomes crucial to leverage the learned information in the pre-trained model for predictions on data in the target distribution. To achieve this, \textit{SOGA}~\cite{mao2024source} preserves the knowledge in the pre-trained model by maximizing the mutual information between the inputs and outputs of the model. In this manner, the discriminate ability of the pre-trained models could be retained and utilized during fine-tuning on the target distribution. Meanwhile, directly fine-tuning the model presents a risk of overfitting to the irrelevant information in data from the target distribution~\cite{suresh2021adversarial}. To avoid capturing irrelevant information that leads to overfitting, \textit{GAPGC} ~\cite{chen2022graphtta} employs the strategy of contrastive learning as a self-supervised task to resist overfitting. GAPGC further designs a trainable GNN-based augmentation module to generate more samples for contrastive learning while decreasing the irreverent information learned by the model. 
Given a model pre-trained on the source distribution, a straightforward strategy for test-time adaptation is to fine-tune the entire model on the target distribution. It is essential to utilize the learned information from the pre-trained model for predictions on data in the target distribution. To achieve this, SOGA \cite{mao2024source} focuses on preserving the knowledge within the pre-trained model by maximizing the mutual information between the inputs and outputs of the model. This approach helps retain and utilize the discriminative ability of the pre-trained models during fine-tuning on the target distribution. However, direct fine-tuning presents a risk of overfitting to irrelevant features in the target distribution's data \cite{suresh2021adversarial}. To mitigate this, GAPGC \cite{chen2022graphtta} adopts contrastive learning as a self-supervised task to prevent overfitting. Furthermore, GAPGC introduces a trainable GNN-based augmentation module that generates additional samples for contrastive learning, thereby reducing the irrelevant information absorbed by the model.

\vspace{0.02in}
\noindent\textbf{Partial Fine-tuning.} 
%Existing strategies have proposed to split the model into two parts to reduce fine-tuning cost: the invariant part and the specific part. Only the specific part is fine-tuned during adaptation, while the invariant part remains fixed. Such a strategy could preserve the legend information in the pre-trained model while also ensuring the adaptation to the target distribution. In \textit{GT3}~\cite{wang2022test} the model consists of two modules: a classification module and a self-supervised module. During updation, the self-supervised module is fine-tuned on the target distribution, while the classification module is fixed. More recently, \textit{GraphGLOW}~\cite{zhao2023graphglow} employs a structural module that is shared across distributions, along with additional GNN layers that are specific to each distribution. By only fine-tuning these additional layers, GraphGLOW achieves adaptation to the target distribution without fine-tuning the entire model.
Existing strategies often propose dividing the model into two components to reduce the costs associated with fine-tuning: an invariant part and a specific part. During adaptation, only the specific part is fine-tuned, while the invariant part remains unchanged. This approach not only preserves the essential information from the pre-trained model but also facilitates adaptation to the target distribution. In GT3 \cite{wang2022test}, the model is split into two modules: a classification module and a self-supervised module. During adaptation, the self-supervised module is fine-tuned on the target distribution, while the classification module remains fixed. More recently, \textit{GraphGLOW} \cite{zhao2023graphglow} uses a structural module that is consistent across distributions, complemented by additional GNN layers that are specific to each distribution. By fine-tuning only these additional layers, GraphGLOW achieves adaptation to the target distribution without the need to fine-tune the entire model.

\begin{comment}
    \subsubsection{Graph Transfer Learning}
Graph transfer learning aims to preserve the performance of graph ML models after training on self-supervised tasks on other new tasks with limited fine-tuning. Different from graph domain adaptation, the objectives in graph transfer learning are typically distinct between training and inference.
%
A popular graph transfer learning strategy is to combine different optimization objectives to enhance model adaptability. For example, AUX-TS~\cite{han2021adaptive} proposes the adaptive auxiliary learning strategy, which aims to accompany target downstream tasks with self-supervised auxiliary tasks during fine-tuning. The authors further consider task similarity to maximally increase the usefulness of each auxiliary task when applied to downstream data. In concrete, these auxiliary tasks can effectively leverage learned information for fine-tuning graph ML models on downstream tasks. 
%

Other works also explore the potential of learning domain-invariant representations that can adapt to different data distributions. For example, AdaGCN~\cite{dai2022graph} proposes an adversarial strategy under semi-supervised learning, where the adversary aims to mitigate the data distribution shift between self-supervised and other tasks to facilitate the transfer of learned knowledge.
\end{comment}